\documentclass[preprint,12pt]{elsarticle}

 \usepackage{booktabs}
\usepackage{amssymb}
\usepackage{amsmath}
\usepackage[strings]{underscore}
\usepackage{algorithm}
\usepackage{pdflscape}
\usepackage{microtype}
\usepackage{adjustbox}
\usepackage{changepage}
\usepackage{algpseudocode}
\usepackage[colorlinks=true, citecolor=blue, linkcolor=blue, urlcolor=blue]{hyperref}

\journal{Nuclear Physics B}

\begin{document}

\begin{frontmatter}



\title{XGRVFL-MV: Residual-Coupled Graph-Embedded Multi-View Random Vector Functional Link Network with FleXi Guardian Loss} %% Article title
\author[inst1]{Yogesh Kumar}
\ead{yogesh.23csz0014@iitrpr.ac.in}

\author[inst1]{Mudasir Ganaie \corref{cor1}}
\ead{mudasir@iitrpr.ac.in}
\cortext[cor1]{Corresponding author}

\affiliation[inst1]{organization={Department of Computer Science and Engineering, Indian Institute of Technology Ropar}, 
            city={Rupnagar},
            postcode={140001}, 
            state={Punjab},
            country={India}}
            
\begin{abstract}

Random Vector Functional Link (RVFL) networks provide an efficient randomized learning framework for classification. Existing multi-view RVFL methods utilize complementary information from multiple views. However, preserving view-specific geometric structure, limiting the influence of large prediction residuals, and modeling relationships between multiple views remain challenging. This paper proposes a Residual-Coupled Graph-Embedded Multi-View RVFL model with fleXi guardian loss (XGRVFL-MV) for multi-view classification. The proposed model constructs RVFL representation for each view, incorporates graph embedding with intrinsic and penalty graphs constructed using the Local Fisher Discriminant Analysis weighting scheme. It also uses the bounded and asymmetric FleXi Guardian (XG) loss for residual learning. A residual-coupling term is introduced to encourage consistency among view-specific prediction residuals while preserving view-specific representations. The resulting optimization problem is solved using an inversion-free first-order optimization procedure based on Nesterov accelerated gradient descent. We evaluate the proposed model on UCI, KEEL, AwA, and Corel5k benchmark datasets. Experimental results, together with statistical analyses and hyperparameter sensitivity analyses, show that XGRVFL-MV achieves competitive classification performance compared with the baseline methods across the evaluated benchmark datasets.

\end{abstract}


% \begin{highlights}
% \end{highlights}

%% Keywords

\end{frontmatter}

\section{Introduction}
Artificial neural networks (ANNs) have been widely used for classification because of their ability to model complex nonlinear relationships \cite{yang2009artificial}. ANNs are generally trained using back-propagation, where network parameters are iteratively updated to minimize prediction errors. However, gradient-descent-based training may suffer from slow convergence \cite{jacobs1988increased}, difficulty in reaching the global minima \cite{gori1992problem}, and sensitivity to the choice of learning rate and parameter initialization. These limitations have motivated the development of alternative learning approaches with simpler and more efficient training procedures.

Randomized neural networks address these difficulties by randomly generating hidden-layer parameters and computing output weights through a closed-form solution \cite{suganthan2021origins, zhang2016survey}. The Extreme Learning Machine (ELM) follows the same randomized learning principle as the Random Vector Functional Link network without direct links (RVFLWoDL), where the hidden-layer parameters are randomly generated and fixed, and the output weights are obtained through a closed-form solution \cite{huang2006extreme}. The random vector functional link (RVFL) network follows randomized learning architecture  includes direct connections from the input layer to the output layer \cite{pao1994learning}. Direct connections allow the output layer to use both the original input features and the nonlinear features generated by the hidden layer. RVFL has universal approximation capability, enabling it to model complex nonlinear relationships \cite{pao1994learning}.

The efficient training of RVFL has led to several extensions that improve its representation and learning capability. Sparse pre-training methods improve the random feature mapping \cite{malik2022extended}, while neuro-fuzzy RVFL combines fuzzy membership functions with randomized hidden nodes \cite{sajid2024neuro}. Kernel-based methods capture nonlinear relationships in a high-dimensional feature space \cite{quadir2025randomized, quadir2025trkm}, whereas graph-embedded RVFL methods preserve the geometric relationships among samples through graph regularization \cite{tanveer2025grvfl}. Deep and ensemble RVFL models use multiple randomized layers or combine several learners to obtain different feature representations \cite{malik2023random}. Despite these developments, most RVFL models are designed for single-view learning and do not explicitly exploit complementary information available from multiple feature representations of the same object.

Single-view setting limits the use of standard RVFL in problems where the same object has multiple feature representations. An image can be represented by colour and texture features, a person by facial and fingerprint features, and a webpage by its text and the anchor text of incoming links \cite{xu2013survey}. Multi-view learning (MVL) uses multiple representations to improve the learning process and finds applications in image annotation, bioinformatics, and social network analysis \cite{berahmand2025comprehensive, qin2025survey}. MVL mainly follows two principles: consensus and complementarity. Consensus encourages agreement among the predictions obtained from different views, whereas complementarity preserves the distinct information provided by each view \cite{xu2013survey}. These principles appear in several multi-view classifiers, including multi-view twin support vector machines \cite{xie2015multi}, parametric-margin models \cite{quadir2024multiview}, multi-weight-vector projection models \cite{yan2025multi}, margin-distribution machines \cite{hu2024multiview}, and structural twin SVMs with safe screening rules \cite{liu2025multi}. Multi-view RVFL also applies consensus and complementarity within a randomized learning framework for DNA-binding protein prediction \cite{quadir2024multiview}. These developments demonstrate that RVFL networks can learn from multiple complementary feature representations within a randomized learning framework.

Although multi-view learning uses complementary information from different feature representations, existing methods ignore the intrinsic geometric structure within each view. Samples that lie close to each other on the underlying data manifold may share similar characteristics, even when their original feature representations differ \cite{belkin2006manifold}. Graph embedding (GE) captures such relationships by constructing an intrinsic graph to preserve similarities among related samples and a penalty graph to separate dissimilar samples \cite{yan2006graph, xu2021understanding}. Local Fisher Discriminant Analysis (LFDA) further uses local similarities to preserve the structure of multimodal classes \cite{sugiyama2007dimensionality}. Graph-embedded multi-view RVFL methods incorporate view-specific graph information into the learning objective through graph regularization and couple the view-specific models to learn from multiple representations \cite{tanveer2025grvfl}. This approach introduces geometric structure preservation into multi-view RVFL learning through graph regularization.

Although graph embedding preserves geometric relationships among samples, it does not explicitly account for noisy samples or outliers during the learning process. Multiview Graph-embedded RVFL methods commonly use the $L_2$ loss, which assigns a large penalty to large residuals. As a result, noisy, mislabelled, or corrupted samples may impact the learned model parameters. Bounded loss functions address this issue by limiting the contribution of large residuals during training \cite{fu2023robust, zhang2024bounded}. Several RVFL models use bounded smooth losses to reduce the influence of extreme residuals and support first-order optimization without matrix inversion \cite{sajid2024wave, akhtar2024advancing}. Asymmetric loss functions have been studied to improve robustness to label noise by assigning different penalties to prediction errors according to their sign \cite{zhou2021asymmetric}. In regression-based learning, this property allows positive and negative prediction residuals to receive different penalties \cite{akhtar2025towards}. Bounded losses also appear in multi-view margin-based classifiers to handle noisy and inconsistent observations across views \cite{quadir2025enhancing, arora2026robust}.

% However, these methods are developed within margin-based learning frameworks and do not consider robust residual learning in graph-embedded multi-view RVFL models.

Despite these advances, a gap remains in graph-embedded multi-view RVFL learning. Existing methods preserve the geometric structure of each view and use cross-view coupling; however, $ L_2$-based optimization remains sensitive to large residuals. Moreover, existing graph-embedded multi-view RVFL methods couple the views through the inner product of view-specific residuals. While this formulation promotes joint learning across the views, it does not explicitly model the discrepancy between their prediction residuals \cite{tanveer2025grvfl}. Existing  RVFL methods uses robust loss functions and inversion-free first-order optimization to improve robustness to large prediction residuals. However, these methods are developed for single-view learning and do not address complementary multi-view information or relationships between prediction residuals across multiple views \cite{akhtar2025towards, sajid2024wave}. Existing graph-embedded multi-view RVFL methods do not simultaneously address geometric structure preservation, robust residual learning, explicit cross-view residual consistency, and inversion-free optimization within a unified learning framework.

To address these limitations, we propose the Residual-Coupled Graph-Embedded Multi-View RVFL with FleXi Guardian (XG) Loss (XGRVFL-MV) framework for multi-view classification.  The model constructs an RVFL representation for each view and preserves the geometric structure of each view through graph regularization. Intrinsic and penalty graphs are constructed using the LFDA weighting scheme to preserve local discriminative relationships and account for multimodal class distributions \cite{sugiyama2007dimensionality}. The model uses the bounded and asymmetric XG loss to reduce the influence of large prediction errors, control residual asymmetry, and bound the loss \cite{akhtar2025towards}. We formulate a residual-consistency mechanism that couples the views by minimizing the discrepancy between their prediction residuals. This formulation promotes agreement between view-specific prediction residuals while preserving separate view-specific representations. The resulting optimization problem is solved using an inversion-free first-order optimization procedure based on Nesterov accelerated gradient descent.
The main highlights of this paper are:
\begin{itemize}
    \item A graph-embedded multi-view Random Vector Functional Link model (XGRVFL-MV) is proposed for multi-view classification by jointly learning view-specific RVFL representations within a unified optimization framework.

    \item A bounded asymmetric fleXi guardian (XG) loss and a residual-coupling mechanism are incorporated into the learning framework to reduce the influence of large prediction residuals while encouraging consistency between the residuals of different views.

    \item  LFDA-based graph embedding is used independently for each view to incorporate intrinsic and penalty graph information through graph regularization, thereby preserving the local geometric relationships within each view.

    \item An inversion-free first-order optimization framework based on Nesterov accelerated gradient descent is formulated to estimate the model parameters without requiring matrix inversion.

    \item Experimental evaluation on UCI, KEEL, AwA, and Corel5k benchmark datasets, together with statistical analyses and hyperparameter sensitivity analyses, demonstrates the effectiveness of the proposed XGRVFL-MV model compared with the baseline methods.

\end{itemize}

This paper is organized as follows.  \autoref{two} reviews the background and related work, including the RVFL network, the graph embedding framework,  the Flexi guardian loss, and existing multi-view learning methods. \autoref{three} presents the proposed XGRVFL-MV model and describes its formulation and optimization. \autoref{four} reports the experimental setup, comparative results, statistical analyses, and hyperparameter sensitivity analyses. Finally, \autoref{five} concludes the paper.

\section{Preliminary Work}
\label{two}
We first introduce the notation used throughout this paper. The subsequent subsections briefly review the RVFL network~\cite{pao1994learning}, the flexi guardian loss~\cite{akhtar2025towards}, the graph embedding framework~\cite{xu2021understanding}, and multi-view learning, which provide the foundation for the proposed XGRVFL-MV model.
\subsection{Notations}

Consider a supervised binary classification problem.
Let the training dataset be defined as
\(\mathcal{S} = \{(x_i^{(1)}, x_i^{(2)}, y_i)\}_{i=1}^{m},\)
where $x_i^{(v)} \in \mathbb{R}^{d_v}$ denotes the feature vector of the
$i$-th sample corresponding to view $v \in \{1,2\}$, and
$y_i \in \{-1, +1\}$ is the associated class label.
For each view, the input data matrix is defined as
\[
X^{(v)} = \left[x_1^{(v)}, x_2^{(v)}, \dots, x_m^{(v)}\right]^\top
\in \mathbb{R}^{m \times d_v}.
\]
Let $Y \in \mathbb{R}^{m \times 1}$ denote the label vector,
where the $i$-th entry corresponds to the class label of the $i$-th sample.

% We consider a two-view supervised classification problem. Let the training dataset be defined as
% \begin{equation}
% \mathcal{S} = \{(x_i^{(1)}, x_i^{(2)}, y_i)\}_{i=1}^{m},
% \end{equation}
% where $x_i^{(v)} \in \mathbb{R}^{d_v}$ denotes the feature vector of the $i$-th sample corresponding to view $v \in \{1,2\}$, and $y_i \in \{1,2,\dots,c\}$ is the associated class label. Let $Y \in \mathbb{R}^{m \times c}$ denote the one-hot encoded label matrix.

\subsection{Random Vector Functional Link Representation}

The random vector functional link (RVFL) network~\cite{pao1994learning} consists of an input layer,
a hidden layer, and an output layer. The original input features are directly
connected to the output layer along with the nonlinear features generated by
the hidden layer. The hidden-layer weights and biases are randomly generated
and remain fixed during training, whereas the output weights are estimated
from the training data.

\noindent
For the $v$-th view, let
$X^{(v)} \in \mathbb{R}^{m \times d_v}$ denote the input feature matrix,
where $m$ is the number of training samples and $d_v$ is the corresponding
feature dimension. The output of the randomized hidden layer is defined as
\begin{equation}
\label{eq:hidden-output}
H^{(v)}
=
\phi\!\left(
X^{(v)}W^{(v)}+\mathbf{e}B^{(v)}
\right),
\end{equation}
where $W^{(v)} \in \mathbb{R}^{d_v \times h_v}$ is the randomly generated
hidden-layer weight matrix, $B^{(v)} \in \mathbb{R}^{1 \times h_v}$ is the
bias vector, $h_v$ denotes the number of hidden nodes, and
$\mathbf{e} \in \mathbb{R}^{m \times 1}$ is a vector of ones. The function
$\phi(\cdot)$ denotes an element-wise nonlinear activation function.

\noindent
Let $x_i^{(v)} \in \mathbb{R}^{1\times d_v}$ denote the $i$-th sample of
the $v$-th view, $w_j^{(v)} \in \mathbb{R}^{d_v\times 1}$ denote the
$j$-th column of $W^{(v)}$, and $b_j^{(v)}$ denote the corresponding bias.
The hidden-layer output, in matrix form, is given as
\begin{equation}
\label{eq:hidden-matrix}
H^{(v)}
=
\begin{bmatrix}
\phi\!\left(x_1^{(v)}w_1^{(v)}+b_1^{(v)}\right)
&
\cdots
&
\phi\!\left(x_1^{(v)}w_{h_v}^{(v)}+b_{h_v}^{(v)}\right)
\\
\vdots
&
\ddots
&
\vdots
\\
\phi\!\left(x_m^{(v)}w_1^{(v)}+b_1^{(v)}\right)
&
\cdots
&
\phi\!\left(x_m^{(v)}w_{h_v}^{(v)}+b_{h_v}^{(v)}\right)
\end{bmatrix}
\in \mathbb{R}^{m\times h_v}.
\end{equation}
\begin{figure}[htbp]
  \centering
\includegraphics[width=0.6\textwidth]{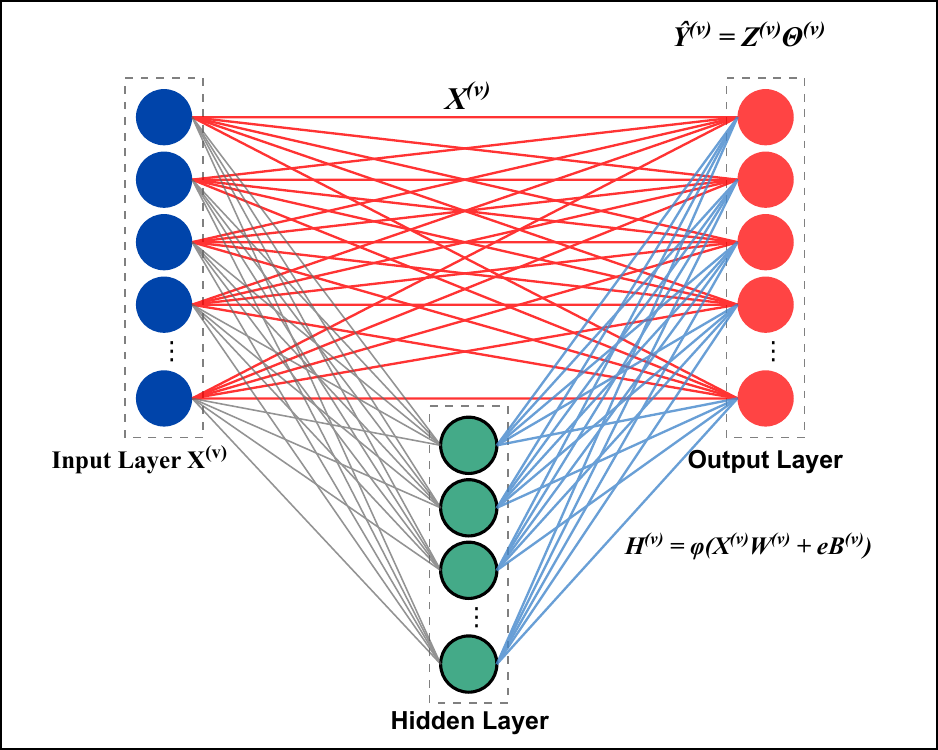}
\caption{Architecture of the RVFL network with a randomized hidden layer and direct connections from the input layer to the output layer.}
  \label{fig:Rvfl}
\end{figure}

\noindent
The original input
features and the hidden-layer output is given as:
\begin{equation}
\label{eq:rvfl-representation}
Z^{(v)}
=
\left[
X^{(v)} \;\; H^{(v)}
\right]
\in
\mathbb{R}^{m\times(d_v+h_v)}.
\end{equation}
For classification, let
$\Theta^{(v)} \in \mathbb{R}^{(d_v+h_v)\times 1}$ denote the output weight
vector associated with the $v$-th view. The corresponding prediction output is given by
\begin{equation}
\label{eq:view-output}
\hat{Y}^{(v)}
=
Z^{(v)}\Theta^{(v)}.
\end{equation}
The standard RVFL model solves the
following optimization problem:
\begin{equation}
\label{eq:rvfl-objective}
\min_{\Theta^{(v)}}
\;
\frac{1}{2}
\left\|
\Theta^{(v)}
\right\|_2^2
+
\frac{c}{2}
\left\|
\xi^{(v)}
\right\|_2^2,
\end{equation}
subject to
\begin{equation}
\label{eq:rvfl-constraint}
Z^{(v)}\Theta^{(v)}-Y
=
\xi^{(v)},
\end{equation}
where $\xi^{(v)}\in\mathbb{R}^{m\times1}$ denotes the prediction residual
and $c>0$ is the regularization parameter.
The closed-form solution of \eqref{eq:rvfl-objective} is given as following:
\begin{equation}
\label{eq:rvfl-closed-form}
\Theta^{(v)}
=
\begin{cases}
\left(
Z^{(v)\top}Z^{(v)}
+
\frac{1}{C}I
\right)^{-1}
Z^{(v)\top}Y,
&
d_v+h_v \leq m,
\\[1.5ex]
Z^{(v)\top}
\left(
Z^{(v)}Z^{(v)\top}
+
\frac{1}{C}I
\right)^{-1}
Y,
&
m < d_v+h_v.
\end{cases}
\end{equation}

\subsection{FleXi Guardian Loss}
\label{subsec:xg_loss_related}

The FleXi Guardian (XG) loss \cite{akhtar2025towards} is a bounded and asymmetric loss function
designed to provide flexible penalization of prediction residuals while
reducing the influence of large errors during learning. Its formulation
allows the shape and upper bound of the loss to be adjusted through two
independent parameters, making it suitable for learning problems where
positive and negative residuals may require different levels of
penalization.
\begin{figure}[h!]
  \centering
\includegraphics[width=\textwidth]{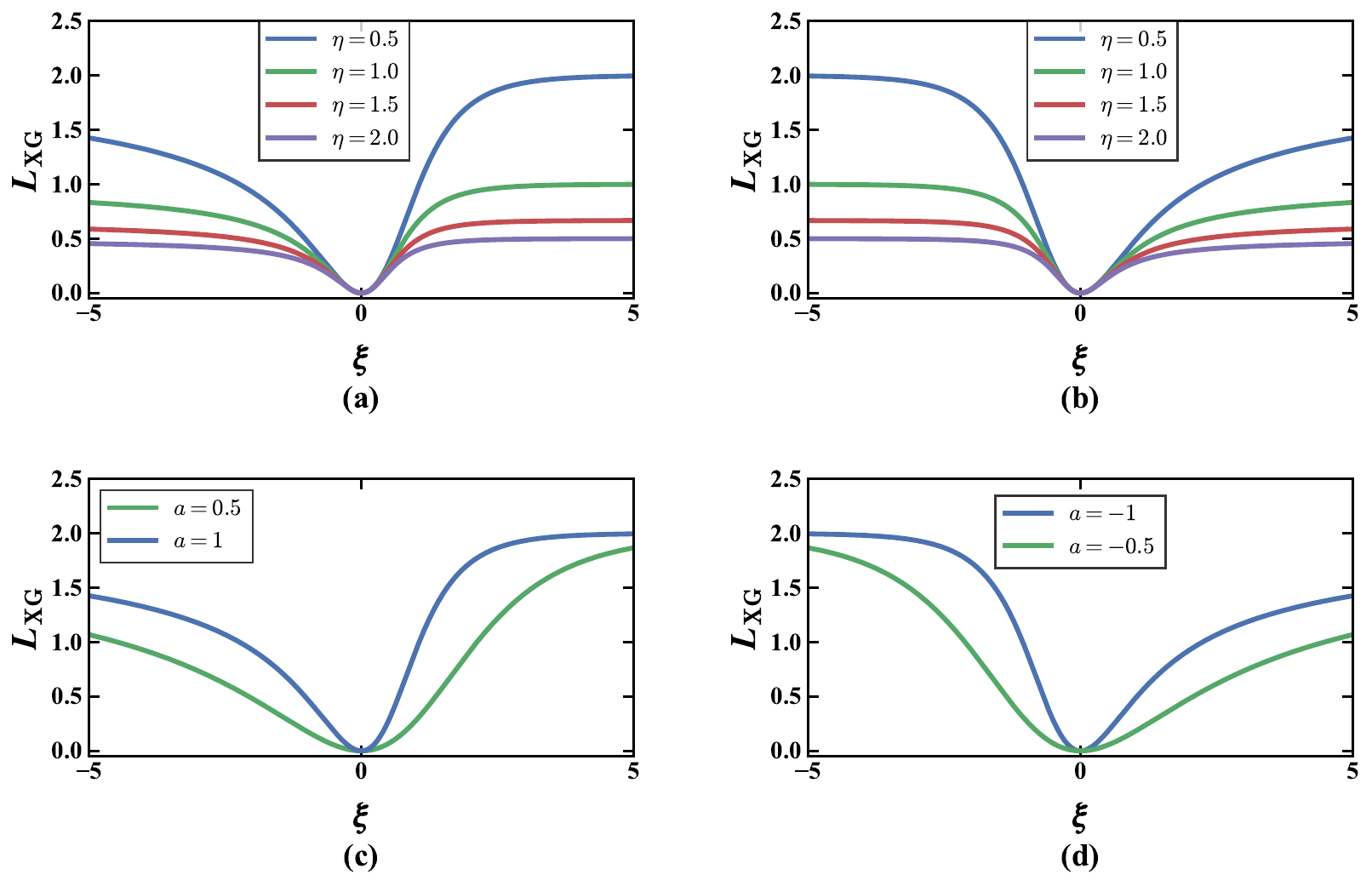}
\caption{Behavior of the XG loss under different parameter settings. Subfigures (a) and (b) illustrate the effect of varying the bounded-control parameter $\eta$ for fixed values of $a=1$ and $a=-1$, respectively. Subfigures (c) and (d) illustrate the effect of varying positive and negative values of $a$, respectively, with $\eta=0.5$. The parameter $a$ controls the asymmetry of the loss, whereas $\eta$ determines its saturation level for large prediction residuals.}
  \label{xgloss}
\end{figure}

\noindent
The XG loss is defined as: 

\begin{equation}
\mathcal{L}_{XG}(\xi)
=
\frac{
a\xi\left(e^{a\xi}-1\right)
}{
1+a\eta\xi\left(e^{a\xi}-1\right)
},
\qquad
\forall\,\xi\in\mathbb{R},
\label{eq:xg_related}
\end{equation}

where $\xi$ denotes the prediction residual, $a\neq0$ controls the
shape and asymmetry of the loss function, and $\eta>0$ determines its upper bound. The formulation in (\ref{eq:xg_related}) consists of two components. The
numerator, $a\xi(e^{a\xi}-1)$, allows the loss to increase smoothly with the prediction residual. The parameter $a$ controls the rate at which the loss increases and also determines the direction of asymmetry. Positive and negative values of $a$ produce different penalization for positive and negative residuals, allowing the loss function to adapt to
different learning requirements.

The denominator, $1+a\eta\xi(e^{a\xi}-1)$, provides a bounded normalization mechanism.
Consequently, the loss gradually approaches a finite upper bound as the magnitude of the residual increases, rather than increasing without limit. The parameter $\eta$ controls this upper bound and provides additional flexibility in adjusting the response of the loss function to large prediction residuals. 
Figure~\ref{xgloss} illustrates the behavior of the XG loss under different parameter settings. Subfigures~(a) and (b) show the effect of varying $\eta$ for fixed values of $a=1$ and $a=-1$, respectively. As $\eta$ increases, the saturation level of the loss decreases while the bounded property is preserved. Subfigures~(c) and (d) illustrate the effect of varying positive and negative values of $a$, respectively, with $\eta=0.5$. Increasing the magnitude of $a$ produces a steeper increase in the loss on the corresponding side of the residual axis, demonstrating its role in controlling the asymmetry of the loss function. The XG loss is continuously differentiable, which enables the use of
gradient-based optimization methods for parameter estimation.

\subsection{Graph Embedding}

Graph embedding (GE) aims to map data samples into a low-dimensional space
while preserving structural and geometric relationships
\cite{xu2021understanding}. In the proposed XGRVFL-MV framework, graph embedding
is performed independently for each view. Let $Z^{(v)} \in \mathbb{R}^{m \times
(d_v+h_v)}$ denote the feature matrix, which is
used to construct the intrinsic graph $G_{\text{int}}^{(v)}$ and the penalty
graph $G_{\text{pen}}^{(v)}$.
The intrinsic and penalty graphs are defined as
$G_{\text{int}}^{(v)} = \{ Z^{(v)}, \Delta_{\text{int}}^{(v)} \}$ and
$G_{\text{pen}}^{(v)} = \{ Z^{(v)}, \Delta_{\text{pen}}^{(v)} \}$, respectively.
Here, $\Delta_{\text{int}}^{(v)} \in \mathbb{R}^{m \times m}$ denotes the
similarity weight matrix that captures pairwise relationships between samples, while $\Delta_{\text{pen}}^{(v)} \in \mathbb{R}^{m \times m}$
denotes the degree matrix associated with the penalty graph.
The graph embedding optimization problem for view $v$ is formulated as follows:
\begin{equation}
\label{eq:graph-embedding}
\begin{aligned}
w^{(v)*}
&= \underset{\operatorname{tr}(w^{(v)\top} Z^{(v)\top} M_{\text{pen}}^{(v)} Z^{(v)} w^{(v)}) = f}
{\operatorname{argmin}}
\sum_{i \neq j}
\left\| w^{(v)\top} z_i^{(v)} - w^{(v)\top} z_j^{(v)} \right\|^2
\Delta_{\text{int},ij}^{(v)} \\
&= \underset{\operatorname{tr}(w^{(v)\top} Z^{(v)\top} M_{\text{pen}}^{(v)} Z^{(v)} w^{(v)}) = f}
{\operatorname{argmin}}
\operatorname{tr}
\!\left(
w^{(v)\top} Z^{(v)\top} M_{\text{int}}^{(v)} Z^{(v)} w^{(v)}
\right),
\end{aligned}
\end{equation}
where $\operatorname{tr}(\cdot)$ denotes the trace operator and $f > 0$ is a
constant that fixes the scale of the embedding and
prevents the trivial solution $w^{(v)} = \mathbf{0}$. Minimizing the objective in
\eqref{eq:graph-embedding} encourages samples with strong intrinsic similarity
to remain close in the embedded space. The constraint normalizes the projected representation and avoids arbitrary
scaling of the projection matrix. The matrix $M_{\text{int}}^{(v)}$ denotes the Laplacian matrix of the intrinsic graph
defined as:
\begin{equation}
\label{eq:M}
M_{\text{int}}^{(v)} = D_{\text{int}}^{(v)} - \Delta_{\text{int}}^{(v)},
\end{equation}
where $D_{\text{int}}^{(v)}$ is a diagonal matrix with entries
$D_{\text{int},ii}^{(v)} = \sum_j \Delta_{\text{int},ij}^{(v)}$.
Similarly, the Laplacian matrix of the penalty graph is given by
\begin{equation}
\label{eq:L}
M_{\text{pen}}^{(v)} = D_{\text{pen}}^{(v)} - \Delta_{\text{pen}}^{(v)} ,
\end{equation}
with $D_{\text{pen},ii}^{(v)} = \sum_j \Delta_{\text{pen},ij}^{(v)}$.
The optimization problem in \eqref{eq:graph-embedding} leads to the generalized
eigenvalue problem
\begin{equation}
G_{\text{int}}^{(v)} w = \lambda G_{\text{pen}}^{(v)} w,
\end{equation}
where
$G_{\text{int}}^{(v)} = Z^{(v)\top} M_{\text{int}}^{(v)} Z^{(v)}$ and
$G_{\text{pen}}^{(v)} = Z^{(v)\top} M_{\text{pen}}^{(v)} Z^{(v)}$.
In existing graph embedding based models, the intrinsic and penalty graphs are
combined through a generalized eigenvalue problem, leading to a transformation
matrix of the form $(G_{\text{pen}}^{(v)})^{-1} G_{\text{int}}^{(v)}$ \cite{yan2006graph}.
The proposed XGRVFL-MV model does not explicitly compute a low-dimensional
embedding. Instead, the intrinsic and penalty graph information is included
in the learning objective through a graph regularization term, following the
principles of manifold regularization \cite{belkin2006manifold}.
For each view, the corresponding graph matrix is defined as:
\begin{equation}
G^{(v)}
=
Z^{(v)\top}
\left(
M_{\mathrm{int}}^{(v)}
-
M_{\mathrm{pen}}^{(v)}
\right)
Z^{(v)}.
\end{equation}
% The matrix $G^{(v)}$
% is used in the view-specific graph regularization term of the XGRVFL-MV objective function. Through this term, the graph information associated with
% each view is considered during the estimation of the output weights.

\subsection{Literature Review on Multi-View Learning}

Multi-view learning (MVL) \cite{berahmand2025comprehensive} considers multiple feature representations of the same samples during model training. Since different views may contain complementary information, learning from multiple representations can improve the quality of the learned model compared with using a single representation. Existing multi-view learning methods generally consider two important characteristics of multi-view data: consistency across different views and complementary information contained in individual views.

Among randomized neural network models, the RVFL neural network has been widely studied because of its efficient architecture. Several studies have extended the RVFL framework to multi-view learning. GRVFL-MV \cite{tanveer2025grvfl} uses graph embedding to preserve the geometric structure of each view during
learning. MvRVFL \cite{quadir2024multiview} separates RVFL models that are learned for individual views, and a residual
coupling term is used to model the relationship between view-specific
predictions. These studies show that the RVFL framework can be adapted
to utilize information from multiple views while maintaining
representations for each view.

Multi-view SVM models have also received
considerable attention. Wave-MvSVM \cite{quadir2025enhancing} and
RoBoSS-MvSVM \cite{arora2026robust} uses bounded loss functions to reduce
the influence of noisy samples and outliers while considering
consistency and complementary information across different views.
Structural information has also been incorporated into multi-view SVM
models. Multi-view structural twin SVM
\cite{liu2025multi} preserves intra-class and inter-class relationships through structural regularization, whereas the multi-view large margin distribution machine \cite{hu2024multiview} considers the distribution of classification margins across different views. Other studies investigate view-specific projection vectors, adaptive margin formulations, and Universum learning for heterogeneous feature representations
\cite{yan2025multi,quadir2024multiview,lou2024multi}.

Deep multi-view learning has also been investigated by using multiple nonlinear transformations for each view to obtain hierarchical feature representations \cite{xie2023deep,jia2025deep}. Alternative
loss functions, structural regularization, and adaptive optimization
strategies have also been explored for multi-view learning
\cite{chen2025multi,he2025new}. In addition, safe-screening techniques have been developed to reduce the computational cost of multi-view SVM optimization by identifying samples that do not
contribute to the final solution
\cite{wang2023safe}. Recent surveys on multi-view learning
\cite{berahmand2025comprehensive} identify robustness to noisy views, preservation of view-specific geometric information, and effective use of complementary information as active research directions. 

\section{Proposed Residual-Coupled Graph-Embedded Multi-View RVFL with FleXi Guardian Loss}
\label{three}

This section presents the proposed Residual-Coupled Graph-Embedded Multi-View RVFL with FleXi Guardian  Loss (XGRVFL-MV) framework for multi-view classification. For each view, an RVFL representation is constructed from the original input features and the features generated by the randomized hidden layer. The output weight matrix is learned for each view, and the corresponding prediction residuals are modeled using the XG loss. The geometric relationships among samples are preserved through view-specific graph embedding based on intrinsic and penalty graphs. The relationship between different views is modeled at the residual level by penalizing the discrepancy between their prediction residuals. The proposed learning formulation jointly considers the prediction residuals, the graph information associated with each view, and the residual relationship between the views. The resulting optimization problem is solved using an inversion-free first-order optimization procedure.

% Let $\{(X^{(v)}, Y)\}_{v=1}^{2}$ denote a two-view training dataset, where
% $X^{(v)} \in \mathbb{R}^{m \times d_v}$ represents the feature matrix and
% $Y \in \mathbb{R}^{m \times c}$ denotes the one-hot encoded label matrix.
% For each view, an RVFL-based predictor with an output weight matrix
% $\Theta^{(v)} \in \mathbb{R}^{(d_v + h_v) \times c}$ is constructed.
\noindent
The optimization problem for XGRVFL-MV is formulated as
\begin{align}
\label{eq:gemv-xgrvfl-binary}
\min_{\Theta^{(1)}, \Theta^{(2)}} \;\;
& \sum_{v=1}^{2}
\Bigg[
\frac{e_v}{2}\|\Theta^{(v)}\|_{2}^{2}
+ \mathcal{L}_{XG}^{(v)}
+ \frac{\theta_{v}}{2}
\operatorname{tr}
\!\left(
\Theta^{(v)\top} G^{(v)} \Theta^{(v)}
\right)
\Bigg] \nonumber\\
&\quad \quad\quad\quad\quad\quad\quad\quad\quad\quad\quad\quad \quad\quad+ \frac{\rho}{2}
\left\|
\xi^{(1)}-\xi^{(2)}
\right\|_2^2,
\end{align}
where the residual for view $v$ is defined as
\begin{equation}
\xi^{(v)} = Z^{(v)} \Theta^{(v)} - Y.
\end{equation}
The loss for each view is computed using the XG loss:
\begin{equation}
\mathcal{L}_{XG}^{(v)} =
\sum_{i=1}^{m} 
\mathcal{L}_{XG}\!\left(\xi_{i}^{(v)}\right),
\end{equation}
where
\begin{equation}
\mathcal{L}_{XG}(\xi)
=
\frac{
a \xi \left(e^{a\xi} - 1\right)
}{
1 + a \eta \xi \left(e^{a\xi} - 1\right)
},
\end{equation}
where $a \neq 0$ controls the asymmetry of the loss, and $\eta > 0$
determines its upper bound.

\noindent
The objective function in \eqref{eq:gemv-xgrvfl-binary} contains four
components. The weight regularization term,
$\frac{e_v}{2}\|\Theta^{(v)}\|_{2}^{2}$, penalizes the magnitude of 
view-specific output weights, where $e_v$ determines the contribution of
this term for the $v$-th view. The data-fitting term,
$\mathcal{L}_{XG}^{(v)}$, measures the prediction residuals of each view
using the XG loss. Since the XG loss is bounded, the
contribution of a residual to the loss remains bounded. Its asymmetric
form also permits different loss values for positive and negative residuals,
depending on the value of the parameter $a$. The graph regularization term,
$\frac{\theta_v}{2}
\operatorname{tr}\left(
\Theta^{(v)\top}G^{(v)}\Theta^{(v)}
\right)$, considers the graph information associated with the $v$-th view.
Here, $G^{(v)}$ is obtained from the intrinsic and penalty graph Laplacians,
and $\theta_v$ determines the contribution of the graph regularization term. The residual coupling term,
$\frac{\rho}{2}
\left\|
\xi^{(1)}-\xi^{(2)}
\right\|_2^2$,
penalizes discrepancies between the prediction residuals of the same samples
across the views. This term encourages agreement between the view-specific
predictors at the residual level while allowing each view to retain its own
RVFL representation and output weight vector. The parameter $\rho$ controls
the contribution of the residual coupling term to the overall objective
function. The objective function components in \eqref{eq:gemv-xgrvfl-binary} that depend on $\Theta^{(v)}$ are differentiated with respect to $\Theta^{(v)}$ as follows.

% \begin{equation}
% \label{eq:objective-decomposition}
% J
% =\;
% \frac{e_v}{2}\|\Theta^{(v)}\|_F^2
% +
% \mathcal{L}_{XG}^{(v)}
% +
% \frac{\theta_v}{2}
% \operatorname{tr}
% \!\left(
% \Theta^{(v)\top}G^{(v)}\Theta^{(v)}
% \right)
% +
% \frac{\rho}{2}
% \|\xi^{(v)}-\xi^{(u)}\|_2^2
% .
% \end{equation}
\noindent
The gradient of the weight regularization component with respect to $\Theta^{(v)}$ is given by
\begin{equation}
\label{eq:gradA}
\nabla_{\Theta^{(v)}}
\left(
\frac{e_v}{2}\|\Theta^{(v)}\|_2^2
\right)
=
e_v\Theta^{(v)}.
\end{equation}
The gradient of the XG-loss component with respect to
$\Theta^{(v)}$ is obtained by applying the chain rule:
\begin{equation}
\label{eq:gradB}
\nabla_{\Theta^{(v)}}\mathcal{L}_{XG}^{(v)}
=
Z^{(v)\top}\Psi^{(v)},
\end{equation}
where $\Psi^{(v)}$ denotes the vector containing the partial derivatives
of the XG loss with respect to the corresponding prediction residuals.
The $i$-th element of $\Psi^{(v)}$ is defined as
\begin{equation}
\label{eq:psi-definition}
\Psi_i^{(v)}
=
\frac{
\partial\mathcal{L}_{XG}\!\left(\xi_i^{(v)}\right)
}{
\partial\xi_i^{(v)}
}.
\end{equation}
Substituting the definition of the XG loss into
(\ref{eq:psi-definition}) yields
\begin{equation}
\label{eq:xg-derivative}
\frac{d\mathcal{L}_{XG}}{d\xi}
=
\frac{
a\left[(a\xi+1)e^{a\xi}-1\right]
}{
\left[
1+a\eta\xi\left(e^{a\xi}-1\right)
\right]^2
}.
\end{equation}
The gradient of the graph regularization component with respect to
$\Theta^{(v)}$ is
\begin{equation}
\label{eq:gradC}
\nabla_{\Theta^{(v)}}
\left[
\frac{\theta_v}{2}
\operatorname{tr}
\!\left(
\Theta^{(v)\top}G^{(v)}\Theta^{(v)}
\right)
\right]
=
\theta_vG^{(v)}\Theta^{(v)}.
\end{equation}
The residual-coupling component contributes the following gradient:
\begin{equation}
\label{eq:gradD}
\nabla_{\Theta^{(v)}}
\left[
\frac{\rho}{2}
\|\xi^{(v)}-\xi^{(u)}\|_2^2
\right]
=
\rho Z^{(v)\top}
\left(
\xi^{(v)}-\xi^{(u)}
\right).
\end{equation}
The overall gradient of the objective function with respect to
$\Theta^{(v)}$ is obtained by summing the contributions of the four
objective function components:
\begin{equation}
\label{eq:final-gradient}
\nabla_{\Theta^{(v)}}J
=
e_v\Theta^{(v)}
+
Z^{(v)\top}\Psi^{(v)}
+
\theta_vG^{(v)}\Theta^{(v)}
+
\rho Z^{(v)\top}
\left(
\xi^{(v)}-\xi^{(u)}
\right),
\qquad u\neq v.
\end{equation}

\subsection{Prediction Rule}

For a new test sample, let $x_{*}^{(v)} \in \mathbb{R}^{d_v}$ denote its
representation in the $v$-th view. The corresponding  RVFL feature
representation is given by
\begin{equation}
\tilde{z}_{*}^{(v)}
=
\left[
x_{*}^{(v)},\;
\phi\!\left(
x_{*}^{(v)}W^{(v)} + B^{(v)}
\right)
\right],
\end{equation}
where $W^{(v)}$ and $B^{(v)}$ denote the randomly generated hidden-layer
weights and biases, respectively, and $\phi(\cdot)$ denotes the activation
function. The hidden-layer parameters used during training remain fixed for
constructing the test representations.
The prediction score for the $v$-th view is defined as
\begin{equation}
s_{*}^{(v)}
=
\tilde{z}_{*}^{(v)}\Theta^{(v)},
\end{equation}
where $\Theta^{(v)}$ denotes the learned output weights of the corresponding
view. Based on the two view-specific scores, three decision functions are
defined as
\begin{align}
\label{decision1}
& P_{c}
=
\operatorname{sign}\!\left[
\frac{1}{2}
\sum_{v=1}^{2}
\tilde{z}_{*}^{(v)}\Theta^{(v)}
\right],\\
\label{decision2}
& P_{1}
=
\operatorname{sign}\!\left(
\tilde{z}_{*}^{(1)}\Theta^{(1)}
\right),\\
\label{decision3}
&P_{2}
=
\operatorname{sign}\!\left(
\tilde{z}_{*}^{(2)}\Theta^{(2)}
\right).
\end{align}
Here, $P_{1}$ and $P_{2}$ denote the predictions obtained from the first and
second views, respectively, while $P_{c}$ denotes the combined prediction
obtained from both views. 

\subsection{Local Fisher Discriminant Analysis under the Graph-Embedded Framework}

For each view, graph information is constructed from the corresponding RVFL
representation $Z^{(v)}$. Two graphs, referred to as the intrinsic graph and
the penalty graph, are considered to describe the relationships among the
training samples. The intrinsic graph represents the relationships between
samples belonging to the same class, whereas the penalty graph considers
both within-class and between-class relationships according to their
respective weighting rules.

The graph weights are defined using the weighting scheme of Local Fisher
Discriminant Analysis (LFDA) \cite{sugiyama2007dimensionality}. For the
$v$-th view, the intrinsic and penalty graph weights are given by
\begin{align}
\Delta_{ij}^{(v),\mathrm{int}}
&=
\begin{cases}
\dfrac{\lambda_{ij}^{(v)}}{N_{c_j}}, & c_i=c_j,\\
0, & \text{otherwise},
\end{cases}
\label{eq:lfda-int}
\\[6pt]
\Delta_{ij}^{(v),\mathrm{pen}}
&=
\begin{cases}
\lambda_{ij}^{(v)}
\left(
\dfrac{1}{m}-\dfrac{1}{N_{c_j}}
\right), & c_i=c_j,\\
\dfrac{1}{m}, & \text{otherwise}.
\end{cases}
\label{eq:lfda-pen}
\end{align}
Here, $N_{c_j}$ denotes the number of training samples belonging to class
$c_j$, and $\lambda_{ij}^{(v)}$ denotes the similarity between samples
$z_i^{(v)}$ and $z_j^{(v)}$. The similarity is computed using the Gaussian
kernel
\begin{equation}
\label{eq:lfda-similarity}
\lambda_{ij}^{(v)}
=
\exp\!\left(
-\frac{
\left\|z_i^{(v)}-z_j^{(v)}\right\|_2^2
}{
2\sigma^2
}
\right),
\end{equation}
where $\sigma>0$ denotes the kernel scaling parameter. The resulting
intrinsic and penalty weights are used to construct the corresponding graph
matrices for each view.

\begin{algorithm}[H]
\caption{Training procedure of the proposed XGRVFL-MV method}
\label{alg:gemv-xgrvfl}
\footnotesize
\begin{algorithmic}[1]

\Require
Training data $\{(X^{(v)},Y)\}_{v=1}^{2}$,
number of hidden nodes $n$,
activation function $\phi(\cdot)$,
parameters $e_v$, $\theta_v$, $\rho$, $a$, and $\eta$,
initial learning rate $\mu_0$,
decay factor $\alpha$,
momentum parameter $\gamma$,
maximum number of iterations $T_{\max}$,
and tolerance $\varepsilon$

\Ensure
Output weight vectors corresponding to multiple views

\For{each view}
    \State Randomly generate $W^{(v)}$ and $B^{(v)}$
    \State Compute $H^{(v)}$ and construct $Z^{(v)}$ using
    \eqref{eq:hidden-output} and \eqref{eq:rvfl-representation}, respectively
    \State Construct $M_{\mathrm{int}}^{(v)}$ and $M_{\mathrm{pen}}^{(v)}$ using
    \eqref{eq:M} and \eqref{eq:L}
    \State Compute
            $G^{(v)}
        =
        Z^{(v)\top}
        \left(
        M_{\mathrm{int}}^{(v)}
        -
        M_{\mathrm{pen}}^{(v)}
        \right)
        Z^{(v)}.$
    \State Initialize
    $\Theta_0^{(v)}=0.01\mathbf{1}$ and
    $V_0^{(v)}=\mathbf{0}$
\EndFor

\For{$t=0,1,\ldots,T_{\max}-1$}
    \State Compute the learning rate
    $\mu_t=\mu_0\alpha^{t/T_{\max}}$

    \For{each view}
        \State Compute the look-ahead point
        $\widetilde{\Theta}_t^{(v)}
        =
        \Theta_t^{(v)}+\gamma V_t^{(v)}$
        \State Compute the corresponding residual
        $\widetilde{\xi}_t^{(v)}
        =
        Z^{(v)}\widetilde{\Theta}_t^{(v)}-Y$
    \EndFor

    \For{each view}
        \State Let $u\neq v$ denote the other view
        \State Compute $\widetilde{\Psi}_t^{(v)}$ element-wise from
        $\widetilde{\xi}_t^{(v)}$ using \eqref{eq:xg-derivative}
        \State Evaluate $g_t^{(v)}=\nabla_{\Theta^{(v)}}J$ at
        $\widetilde{\Theta}_t^{(v)}$ using
        \eqref{eq:final-gradient}, with residuals
        $\widetilde{\xi}_t^{(v)}$ and
        $\widetilde{\xi}_t^{(u)}$
    \EndFor

    \For{each view}
        \State Update the velocity as
        $V_{t+1}^{(v)}
        =
        \gamma V_t^{(v)}-\mu_t g_t^{(v)}$
        \State Update the output weights as
        $\Theta_{t+1}^{(v)}
        =
        \Theta_t^{(v)}+V_{t+1}^{(v)}$
    \EndFor

    \If{$
    \max\limits_{v\in\{1,2\}}
    \left\|
    \Theta_{t+1}^{(v)}-\Theta_t^{(v)}
    \right\|_2
    \leq\varepsilon
    $}
        \State \textbf{break}
    \EndIf
\EndFor

\State \Return $\Theta^{(1)}$ and $\Theta^{(2)}$

\end{algorithmic}
\end{algorithm}

\section{Experiments and Results Discussion}
\label{four}

This section describes the experimental evaluation of the proposed XGRVFL-MV model. The experimental setup is first presented, including the datasets obtained from the UCI~\cite{asuncion2007uci} and KEEL~\cite{derrac2015keel} repositories, the evaluation protocol, and the parameter settings. The performance of the proposed XGRVFL-MV model is 
compared with several existing single-view and multi-view learning methods. Statistical analyses using the Friedman test, Wilcoxon signed-rank test, and win-tie-loss analysis are used to assess model performance.
Finally, the influence of the model parameters is examined through a
parameter sensitivity analysis. The proposed XGRVFL-MV model is compared with baseline methods,
namely RVFLWoDL1~\cite{huang2006extreme} (RVFLWoDL using the first
view), RVFLWoDL2~\cite{huang2006extreme} (RVFLWoDL using the second
view), RVFL1~\cite{pao1994learning} (RVFL using the first view),
RVFL2~\cite{pao1994learning} (RVFL using the second view),
MVLDM~\cite{hu2024multiview},
MvRVFL~\cite{quadir2024multiview}, and
GRVFL-MV~\cite{tanveer2025grvfl}. The comparative results are discussed
in the following subsections.

\subsection{Experimental Setup}

All experiments are conducted on an Intel Xeon(R) W5-2565X processor running at 4.7~GHz with 36 CPU cores, Ubuntu 22.04.4 LTS, and Python~3.9. For each dataset, the samples are randomly divided into 70\% training and 30\% testing sets. Hyperparameters are selected using
five-fold cross-validation on the training set.
For all RVFL-based methods, including the proposed XGRVFL-MV model, the number of hidden neurons is selected from the range $[3,203]$ with a
step size of 20. The regularization parameter is chosen from $\{10^{-4},10^{-3},\ldots,10^{4}\}$. For RVFLWoDL2 and RVFL2, Principal Component Analysis (PCA) is applied to construct the second view.
For the MVLDM model~\cite{hu2024multiview}, the parameters $d_1$, $d_2$, and $d_3$ are assigned the value selected from
$\{1,10,50,100\}$. The parameters
$\lambda_1$, $\lambda_2$, $v_1$, $v_2$, and $\theta$
are selected from $\{0.01,0.1,1,10,100\}$. The penalty parameter $\epsilon$ and the maximum number of ADMM iterations are chosen as $0.01$ and $50$, respectively.
For the GRVFL-MV model~\cite{tanveer2025grvfl}, the parameters are chosen as $d_1=d_2=d_3=d$ and
$\gamma_1=\gamma_2=\gamma$, where both
$d$ and $\gamma$, together with $\mu$, are selected from
$\{10^{-4},10^{-3},\ldots,10^{4}\}$. For the proposed XGRVFL-MV model, the XG-loss parameters
$a$ and $\eta$ are selected from
$\{-1.5, -1, -0.5, 0.5, 1.5\}$ and
$\{0.5, 1.0, 1.5, 2.0\}$, respectively.
% For the AwA dataset, a reduced search space is adopted because of the
% high feature dimensionality. The number of hidden neurons is selected from $[3,64]$ with a step size of 10, and the regularization parameter
% is selected from $\{10^{-3},10^{-2},\ldots,10^{3}\}$.

\subsection{Description of UCI and KEEL Datasets}

The experimental evaluation is conducted on benchmark datasets obtained from the UCI Machine Learning Repository \cite{asuncion2007uci} and the KEEL repository
\cite{derrac2015keel}. Both repositories provide benchmark datasets that are widely used for evaluating machine learning algorithms. The selected datasets cover a range of application domains and differ in the number of samples, feature dimensions, and classification characteristics, providing a diverse collection of benchmark problems for experimental evaluation.

For each dataset, the samples are randomly divided into training and testing sets using a 70\%-30\% split. The resulting training size,
testing size, and number of features for each dataset are summarized in Table~\ref{dataset_description}. The selected datasets include problems with different feature
dimensions, ranging from 2 features (Ripley) to 100 features (Hill-Valley). Most datasets contain between 4 and 33 features and vary considerably in the number of training and testing samples. This variation provides an opportunity to examine the behaviour of the
proposed XGRVFL-MV model under different dataset characteristics while using the same experimental protocol for all benchmark datasets.

\begin{table}[!h]
\centering
\caption{Description of the UCI and KEEL benchmark datasets.}
\label{dataset_description}
\resizebox{0.7\textwidth}{!}{
\begin{tabular}{lccc}
\toprule
\textbf{Dataset} & \textbf{Training} & \textbf{Testing} & \textbf{Features} \\
\midrule
credit-approval                & 483  & 207 & 15  \\
hepatitis                      & 108  & 47  & 19  \\
yeast5                         & 1038 & 446 & 8   \\
vertebral-column-2classes      & 217  & 93  & 6   \\
monks-1                        & 86   & 38  & 6   \\
vehicle1                       & 592  & 254 & 18  \\
parkinsons                     & 136  & 59  & 22  \\
hill-valley                    & 424  & 182 & 100 \\
breast-cancer-wisc-prog        & 138  & 60  & 33  \\
yeast-0-5-6-7-9\_vs\_4         & 369  & 159 & 8   \\
yeast3                         & 1038 & 446 & 8   \\
cmc                            & 1031 & 442 & 9   \\
mammographic                   & 672  & 289 & 5   \\
blood                          & 523  & 225 & 4   \\
pima                           & 537  & 231 & 8   \\
breast-cancer-wisc-diag        & 398  & 171 & 30  \\
ripley                         & 875  & 375 & 2   \\
breast-cancer-wisc             & 489  & 210 & 9   \\
\bottomrule
\end{tabular}
}
\end{table}

\subsection{Results on UCI and KEEL Datasets}

The UCI and KEEL benchmark datasets provide a single feature representation for each sample. To enable multi-view learning, a second view is constructed from the original features following~\cite{arora2026robust}. The original feature set is treated as view~${v_1}$, whereas the PCA-transformed feature set is considered as view~${v_2}$. Following the experimental protocol in~\cite{tanveer2025grvfl}, the principal components are retained to preserve 95\% of the data variance. 

\begin{landscape}
\begin{table*}[t]
\vspace*{-90pt}
\begin{adjustwidth}{-0.2cm}{-0.5cm} 
\caption{Performance comparison of baseline and the proposed XGRVFL-MV model on  UCI and KEEL datasets}
\centering
\label{table:results1}
\setlength{\tabcolsep}{8pt} 
\renewcommand{\arraystretch}{1.7} 
\scalebox{0.53}{%
\begin{tabular}{|c|c|c|c|c|c|c|c|c|}
\hline
  & \textbf{RVFLWoDL1~\cite{huang2006extreme}} & \textbf{RVFLWoDL2~\cite{huang2006extreme}} & \textbf{RVFL1~\cite{pao1994learning}} & \textbf{RVFL2~\cite{pao1994learning}} & \textbf{MV-LDM~\cite{hu2024multiview}}& \textbf{MVRVFL~\cite{quadir2024multiview}} & \textbf{GRVFL-MV~\cite{tanveer2025grvfl}} & \textbf{XGRVFL-MV$^*$} \\\hline
\textbf{Dataset} & \textbf{AUC $\uparrow$} & \textbf{AUC $\uparrow$} & \textbf{AUC $\uparrow$} & \textbf{AUC $\uparrow$} & \textbf{AUC $\uparrow$} & \textbf{AUC $\uparrow$} & \textbf{AUC $\uparrow$} & \textbf{AUC $\uparrow$} \\
 $(Patterns \times Features)$ &  $(n, e)$ & $(n, e)$ & $(n, e)$ & $(n, e)$  & $(e, \epsilon,\lambda, v1, v2, \theta)$ & $(n, e1, e3, \rho)$ & $(n, e, \mu, \gamma)$ & $(n, e, \rho, \theta, a, \eta)$ \\\hline

credit-approval & 83.913 & 83.804 & 83.478 & 83.092 & 48.925 & 69.348 & 83.370 & \textbf{85.109} \\
$(690\times15)$ & $(123,10^{-2})$ & $(163,10^{-3})$ & $(3,10^{2})$ & $(163,10^{-3})$ & $(50,10^{-2},10^{-2},10^{-2},1,10^{-2})$ & $(183,10^{4},10^{-4},10^{4})$ & $(63,10^{-3},10^{-5},10^{-3})$ & $(3,10^{-4},10^{-4},10^{-4},-0.5,0.5)$ \\\hline

hepatitis & 62.297 & 70.946 & 67.297 & 68.649 & 68.649 & 83.243 & 71.892 & \textbf{84.595} \\
$(155\times19)$ & $(123,10^{-3})$ & $(83,10^{-3})$ & $(123,10^{-3})$ & $(63,10^{-3})$ & $(1,10^{-2},2,1,1,1)$ & $(43,10^{-4},10^{-4},10^{-4})$ & $(43,10,10,10^{2})$ & $(83,10^{3},10^{-4},10^{-4},1,0.5)$ \\\hline

yeast5 & 76.808 & \textbf{92.077} & 80.654 & 80.654 & 84.038 & 69.115 & 73.077 & 88.231 \\
$(1484\times8)$ & $(183,10^{2})$ & $(203,10^{2})$ & $(163,10^{2})$ & $(163,10^{2})$ & $(10,10^{-2},-2,2,-2,0)$ & $(183,10^{3},10^{-4},10^{3})$ & $(183,10^{-2},10^{-4},10^{-2})$ & $(163,10^{-2},10^{-2},10^{-3},-0.5,0.5)$ \\\hline

vertebral-column-2clases & 80.317 & 79.524 & 80.317 & 75.476 & 81.191 & 61.429 & 81.984 & \textbf{85.238} \\
$(310\times6)$ & $(183,10^{-2})$ & $(83,10^{-2})$ & $(183,10^{-2})$ & $(183,10^{-3})$ & $(100,10^{-2},-2,2,1,1)$ & $(183,10^{-4},10^{-4},10^{4})$ & $(43,10^{-1},10^{-2},10^{-5})$ & $(123,10^{-4},10^{-4},10^{-3},0.5,0.5)$ \\\hline

monks-1 & 73.684 & 73.684 & 73.684 & 73.684 & 86.842 & \textbf{89.474} & 81.579 & 76.316 \\
$(124\times6)$ & $(163,10^{-2})$ & $(183,10^{-2})$ & $(163,10^{-2})$ & $(183,10^{-2})$ & $(50,10^{-2},-2,1,-2,1)$ & $(103,10^{-4},10^{-4},10^{-4})$ & $(63,10,10^{-5},10^{-5})$ & $(163,10^{4},10^{4},10^{3},1,2)$ \\\hline

vehicle1 & 75.767 & 66.801 & 75.983 & 68.315 & 74.998 & 65.763 & 69.927 & \textbf{81.294} \\
$(846\times18)$ & $(123,10^{2})$ & $(63,10^{2})$ & $(123,10^{3})$ & $(63,10^{4})$ & $(10,10^{-2},-2,0,-2,1)$ & $(183,10^{2},10^{-4},10^{2})$ & $(83,10^{-3},10^{-3},10^{-4})$ & $(183,10^{-4},10^{-4},10^{-4},-0.5,0.5)$ \\\hline

parkinsons & 84.394 & 73.106 & 76.591 & 76.667 & \textbf{86.667} & 85.455 & 85.530 & 83.182 \\
$(195\times22)$ & $(43,10^{2})$ & $(183,10^{3})$ & $(183,10^{-4})$ & $(43,10^{-4})$ & $(1,10^{-2},-2,-2,-2,0)$ & $(123,10^{-4},10^{-4},10^{-4})$ & $(163,10^{-3},10^{-5},10^{-5})$ & $(163,10^{-4},10^{-4},10^{-3},0.5,1)$ \\\hline

hill-valley & 43.816 & 51.377 & 38.841 & 51.377 & \textbf{53.019} & 46.570 & 46.498 & 52.935 \\
$(606\times100)$ & $(63,10^{2})$ & $(23,10^{3})$ & $(63,10^{2})$ & $(23,10^{3})$ & $(100,10^{-2},-2,-2,-1,0)$ & $(183,10^{4},10^{-4},10^{4})$ & $(183,10^{3},10^{4},10^{-4})$ & $(103,10^{-3},10^{3},10^{2},-0.5,0.5)$ \\\hline

breast-cancer-wisc-prog & 66.770 & 63.199 & 65.683 & 63.199 & 61.025 & 56.988 & 58.851 & \textbf{70.652} \\
$(198\times33)$ & $(103,10^{-2})$ & $(143,10^{-3})$ & $(103,10^{-2})$ & $(143,10^{-3})$ & $(10,10^{-2},1,2,-2,2)$ & $(123,10^{-4},10^{-4},10^{-4})$ & $(23,10^{-1},10^{-5},10^{-2})$ & $(43,10^{-4},10^{-4},10^{-3},-1,1.5)$ \\\hline

yeast-0-5-6-7-9\_vs\_4 & 59.306 & 48.264 & 54.931 & 54.931 & \textbf{60.556} & 58.681 & \textbf{60.556} & 57.917 \\
$(528\times8)$ & $(103,10^{2})$ & $(23,10^{2})$ & $(43,10^{2})$ & $(103,10^{2})$ & $(10,10^{-2},-2,1,-2,1)$ & $(143,10^{2},10^{-4},10^{2})$ & $(43,10^{2},10^{-4},10^{4})$ & $(163,10^{-4},10^{-2},10^{-3},-1,0.5)$ \\\hline

yeast3 & 85.223 & 86.370 & 83.434 & 86.370 & 89.305 & 82.679 & 87.390 & \textbf{90.451} \\
$(1484\times8)$ & $(63,10^{2})$ & $(203,10^{-2})$ & $(203,10^{-2})$ & $(203,10^{-2})$ & $(10,10^{-2},1,-2,-2,2)$ & $(183,10^{2},10^{-4},10^{2})$ & $(43,10^{3},10^{2},10^{3})$ & $(83,10^{-4},10^{-2},10^{-4},-1,0.5)$ \\\hline

cmc & 69.344 & 70.067 & 69.344 & 70.067 & 69.721 & 64.071 & \textbf{72.314} & 71.265 \\
$(1473\times9)$ & $(183,10^{-2})$ & $(143,10^{-2})$ & $(183,10^{-2})$ & $(143,10^{-2})$ & $(10,10^{-2},2,-2,2,-1)$ & $(183,10^{4},10^{3},10^{4})$ & $(63,10^{-2},10^{-3},10^{-2})$ & $(183,10^{-4},10^{-4},10^{-3},1,1)$ \\\hline

mammographic & 80.007 & 80.279 & 80.007 & 81.196 & 82.203 & 75.258 & 80.823 & \textbf{83.050} \\
$(961\times5)$ & $(123,10^{-2})$ & $(43,10^{-2})$ & $(123,10^{-2})$ & $(83,10^{2})$ & $(50,10^{-2},-2,1,-2,2)$ & $(105,10^{3},1,10^{3})$ & $(83,10^{3},10^{2},1)$ & $(123,10^{-4},10^{-4},10^{-4},-0.5,2)$ \\\hline

blood & 65.058 & 65.984 & 62.914 & 66.228 & 68.421 & 65.140 & \textbf{69.347} & 68.470 \\
$(748\times4)$ & $(63,10^{2})$ & $(203,10^{-2})$ & $(43,10^{3})$ & $(203,10^{-2})$ & $(50,10^{-2},-2,-2,2,0)$ & $(105,10^{4},10^{-4},10^{4})$ & $(163,10^{-1},10^{-2},10^{-5})$ & $(143,10^{-4},10^{-4},10^{-4},-1,0.5)$ \\\hline

pima & 69.975 & 68.642 & 69.975 & 68.642 & 69.642 & 67.296 & \textbf{73.864} & 70.025 \\
$(768\times8)$ & $(103,10^{-2})$ & $(103,10^{-2})$ & $(103,10^{-2})$ & $(103,10^{-2})$ & $(1,10^{-2},2,-2,-2,1)$ & $(183,10^{3},10^{-4},10^{3})$ & $(163,10^{-3},10^{-4},10^{-4})$ & $(63,10^{-4},10^{-4},10^{-3},0.5,2)$ \\\hline

breast-cancer-wisc-diag & 94.531 & 95.313 & 94.531 & 95.626 & 96.875 & 94.230 & \textbf{98.284} & 96.415 \\
$(569\times30)$ & $(23,10^{-4})$ & $(203,10^{-3})$ & $(23,10^{-4})$ & $(143,10^{-3})$ & $(1,10^{-2},2,-2,-2,1)$ & $(103,10^{2},10^{-4},10^{2})$ & $(23,10^{-4},10^{-5},10^{-1})$ & $(23,10^{2},10^{-4},10^{-2},1,0.5)$ \\\hline

ripley & 90.667 & \textbf{92.266} & 90.667 & \textbf{92.266} & 92.003 & 89.332 & 91.469 & 90.929 \\
$(1250\times2)$ & $(83,10^{4})$ & $(43,10^{2})$ & $(83,10^{4})$ & $(43,10^{2})$ & $(100,10^{-2},-2,2,2,0)$ & $(183,10^{4},10^{-3},10^{4})$ & $(23,10^{-2},10^{-5},10)$ & $(183,10^{-4},10^{-2},10^{-4},1,1.5)$ \\\hline

breast-cancer-wisc & 94.354 & 93.659 & 94.354 & 95.048 & 94.716 & 94.851 & 94.716 & \textbf{95.411} \\
$(699\times9)$ & $(163,10^{-2})$ & $(163,10^{-2})$ & $(163,10^{-2})$ & $(3,10^{-4})$ & $(1,10^{-2},2,2,2,0)$ & $(95,10^{2},10^{-4},10^{2})$ & $(3,10^{-4},10^{-5},10^{-5})$ & $(3,10^{2},10^{3},10^{3},-1,2)$ \\\hline

 \textbf{Average AUC $\uparrow$} & 75.350 & 75.300 & 74.590 & 75.080 & 76.040 & 73.270 & 76.750 & \textbf{79.530} \\ \hline
\textbf{Average Rank $\downarrow$} & 5.28 & 5.03 & 5.78 & 4.69 & 3.42 & 6.17 & 3.42 & \textbf{2.22} \\
\hline
\multicolumn{9}{l}{Here, $^*$ denotes the proposed model.}\\
\end{tabular}
}
\end{adjustwidth}
\end{table*}
\end{landscape}

The proposed XGRVFL-MV model is compared with baseline methods, namely RVFLWoDL1, RVFLWoDL2, RVFL1, RVFL2, MVLDM, MvRVFL, and GRVFL-MV. Table~\ref{table:results1} reports the AUC values obtained by all methods on the benchmark datasets, together with the average AUC and the corresponding average rank.

XGRVFL-MV records the highest AUC among the compared methods on the credit-approval (85.109\%), hepatitis (84.595\%), vertebral-column-2clases (85.238\%), vehicle1 (81.294\%), breast-cancer-wisc-prog (70.652\%), yeast3 (90.451\%), mammographic (83.050\%), and breast-cancer-wisc (95.411\%) datasets.
XGRVFL-MV records an AUC of 83.182\% compared with 86.667\% achieved by MV-LDM on parkinsons dataset, while on breast-cancer-wisc-diag it obtains 96.415\% compared with 98.284\% obtained by GRVFL-MV. Overall, the proposed model remains competitive across most of the benchmark
datasets. Several datasets contain relatively few training samples compared with their feature dimensions, including hepatitis (155 samples, 19 features), breast-cancer-wisc-prog (198 samples, 33 features), and
credit-approval (690 samples, 15 features). On these datasets, RVFLWoDL and RVFL variants generally produce lower AUC values. With respect to hepatitis dataset, RVFLWoDL1 and RVFL1 achieve 62.297\% and 67.297\%, respectively, whereas XGRVFL-MV records the
highest AUC of 84.595\%. 

The results on the imbalanced datasets are more varied. On yeast-0-5-6-7-9\_vs\_4, MV-LDM and GRVFL-MV both achieve an AUC of 60.556\%, while XGRVFL-MV obtains 57.917\%. On the cmc dataset, GRVFL-MV records the highest AUC of 72.314\%, followed closely by
XGRVFL-MV with 71.265\%. In contrast, XGRVFL-MV achieves 88.231\% on yeast5 and the highest AUC of 90.451\% on yeast3. These results show that the relative performance of the competing methods varies across different imbalanced datasets. On monks-1, hill-valley, and blood, the best AUC values are obtained by MvRVFL (89.474\%), MV-LDM (53.019\%), and GRVFL-MV (69.347\%),
respectively. On the ripley dataset, RVFLWoDL2 and RVFL2 both obtain the highest AUC of 92.266\%, while XGRVFL-MV records 90.929\%. Although XGRVFL-MV is not the top-performing method on these datasets, its AUC
remains comparable to those of the strongest baseline methods.

Across all datasets, XGRVFL-MV achieves the highest average AUC of 79.530\%, followed by GRVFL-MV with 76.750\% and MV-LDM with 76.664\%. The proposed XGRVFL-MV model also attains the lowest average rank of
2.22, followed by GRVFL-MV and MV-LDM, both with an average rank of 3.42, whereas MvRVFL records an average rank of 6.17. These results indicate that XGRVFL-MV consistently ranks among the better-performing
methods across the benchmark datasets considered in this study.

\subsection{Results on Animals with Attributes Dataset}

The Animals with Attributes (AwA5) dataset~\cite{lampert2013attribute} contains 30,475 images belonging to 50 animal classes, with each image
represented by six types of pre-extracted features. Following the experimental setting of~\cite{tanveer2025grvfl}, the chimpanzee, giant
panda, leopard, Persian cat, pig, hippopotamus, humpback whale, raccoon, and rat classes are selected for evaluation. Two feature representations are considered for constructing the multi-view data. The first view (View~1) consists of a 252-dimensional Histogram of
Oriented Gradients (HOG) descriptor, whereas the second view (View~2) is represented by a 2000-dimensional $L_{1}$-normalized Speeded-Up Robust Features (SURF) descriptor. 

\begin{figure}[H]
  \centering
\includegraphics[width=0.7\textwidth]{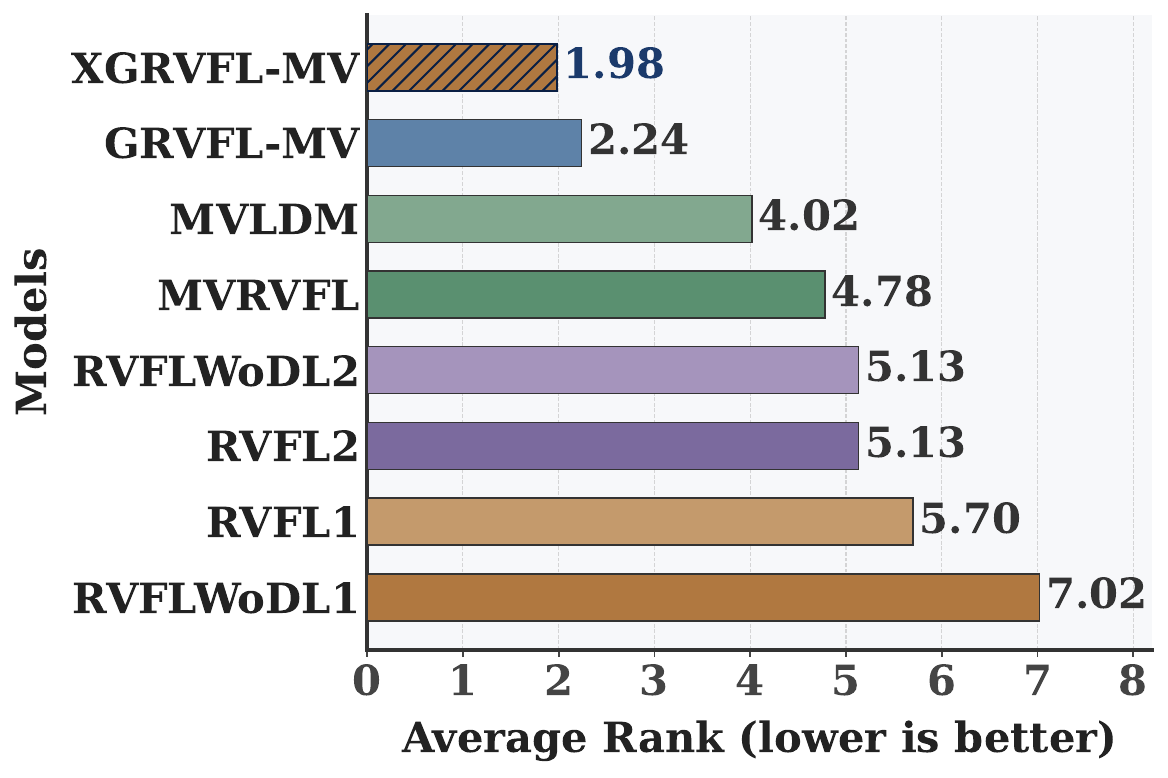}
\caption{Average ranks of the baseline methods and the proposed XGRVFL-MV on the AwA benchmark datasets. Lower ranks indicate better performance.}
  \label{fig:rank}
\end{figure}

\begin{figure}[H]
  \centering
\includegraphics[width=0.7\textwidth]{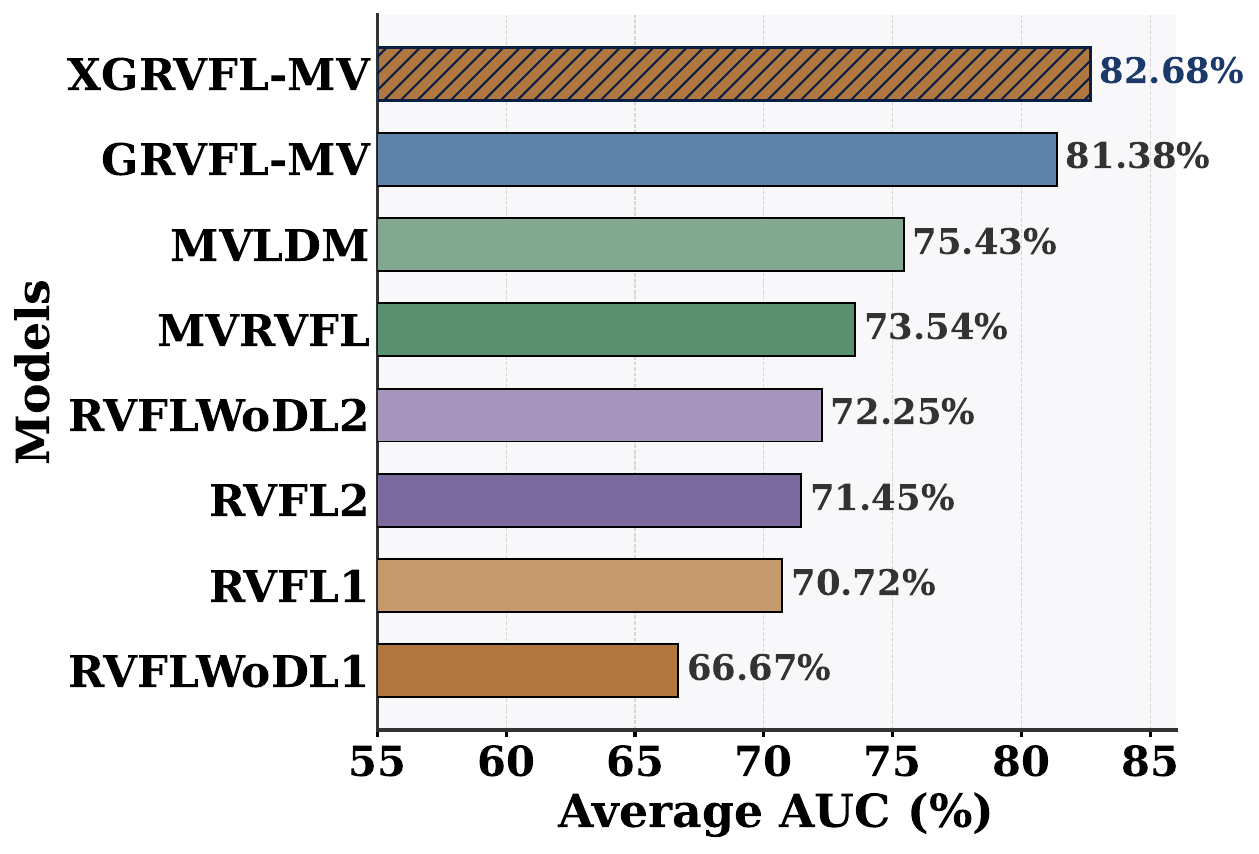}
\caption{Average AUC of the baseline methods and the proposed XGRVFL-MV on the AwA benchmark datasets.}
  \label{fig:AUC}
\end{figure}

Table~\ref{table:Awa} reports the AUC (\%) obtained by the proposed XGRVFL-MV model and baseline methods, namely RVFLWoDL1, RVFLWoDL2, RVFL1, RVFL2, MV-LDM, MvRVFL, and GRVFL-MV, on  classification tasks of the AwA dataset. The corresponding average AUC values and average ranks are illustrated in Figures~\ref{fig:AUC} and~\ref{fig:rank}, respectively. Overall, XGRVFL-MV achieves the highest average AUC of 82.680\% and the lowest average rank of 1.98, followed by GRVFL-MV with an average AUC of 81.380\% and an average rank of 2.24. On the classification tasks, XGRVFL-MV achieves the highest
AUC on Chimpanzee\_vs\_Giantpanda (91.667\%),
Chimpanzee\_vs\_Hippopotamus (85.000\%),
Chimpanzee\_vs\_Leopard $(86.667\%)$,
Chimpanzee\_vs\_Pig (78.333\%),
Persiancat\_vs\_Hippopotamus (88.333\%)
Pig\_vs\_Hippopotamus (66.667\%),
Raccoon\_vs\_Humpback (96.667\%), and
Raccoon\_vs\_Rat (85.000\%). In addition, it shares the highest AUC with
GRVFL-MV on Chimpanzee\_vs\_Raccoon (80.000\%),
Leopard\_vs\_Persiancat (90.000\%), and
Leopard\_vs\_Pig (78.333\%), and with RVFLWoDL2 on
Persiancat\_vs\_Humpbackwhale (95.000\%). For the remaining classification tasks, the proposed XGRVFL-MV model generally produces AUC values close to those of the best-performing method.
Proposed XGRVFL-MV records 85.000\% on
Chimpanzee\_vs\_Persiancat compared with 90.000\% obtained by GRVFL-MV, and 71.667\% on Chimpanzee\_vs\_Rat compared with 81.667\% obtained by MvRVFL. Similarly, on Giantpanda\_vs\_Hippopotamus and
Giantpanda\_vs\_Humpbackwhale, GRVFL-MV achieves
93.333\% and 95.000\%, whereas XGRVFL-MV records
86.667\% and 93.333\%, respectively.

\begin{landscape}
\begin{table*}[t]
\vspace*{-110pt}
\begin{adjustwidth}{-0.2cm}{-0.5cm}
\caption{Performance comparison of baseline and the proposed XGRVFL-MV model on  AwA datasets}
\centering
\label{table:Awa}
\setlength{\tabcolsep}{10pt}
\renewcommand{\arraystretch}{1.5}
\scalebox{0.50}{%
\begin{tabular}{|c|c|c|c|c|c|c|c|c|}
\hline
  & \textbf{RVFLWoDL1~\cite{huang2006extreme}} & \textbf{RVFLWoDL2~\cite{huang2006extreme}} & \textbf{RVFL1~\cite{pao1994learning}} & \textbf{RVFL2~\cite{pao1994learning}} & \textbf{MV-LDM~\cite{hu2024multiview}}& \textbf{MVRVFL~\cite{quadir2024multiview}} & \textbf{GRVFL-MV~\cite{tanveer2025grvfl}} & \textbf{XGRVFL-MV$^*$} \\\hline
\textbf{Dataset} & \textbf{AUC $\uparrow$} & \textbf{AUC $\uparrow$} & \textbf{AUC $\uparrow$} & \textbf{AUC $\uparrow$} & \textbf{AUC $\uparrow$} & \textbf{AUC $\uparrow$} & \textbf{AUC $\uparrow$} & \textbf{AUC $\uparrow$} \\
 $(Patterns \times Features)$ &  $(n, e)$ & $(n, e)$ & $(n, e)$ & $(n, e)$  & $(e, \epsilon,\lambda, v1, v2, \theta)$ & $(n, e1, e3, \rho)$ & $(n, e, \mu, \gamma)$ & $(n, e, \rho, \theta, a, \eta)$ \\\hline

Chimpanzee\_vs\_Giantpanda & 73.333 & 68.333 & 68.333 & 75.000 & 73.333 & 61.667 & 80.000 & \textbf{91.667} \\
$\big(200 \times(131\ \&\ 96)\big)$ & $(23,10^{-3})$ & $(3,10^{-3})$ & $(3,10^{-3})$ & $(43,10^{-3})$ & $(50,10^{-2},2,-2,1,2)$ & $(33,10^{2},10^{-3},10^{2})$ & $(13,1,1,10^{-3})$ & $(23,10,10^{-3},10^{-3},0.5,2)$ \\\hline

Chimpanzee\_vs\_Hippopotamus & 70.000 & 78.333 & 75.000 & 76.667 & 81.667 & 61.667 & 75.000 & \textbf{85.000} \\
$\big(200 \times(131\ \&\ 96)\big)$ & $(23,10^{-1})$ & $(13,10^{-3})$ & $(23,10^{-3})$ & $(13,10^{-2})$ & $(1,10^{-2},-2,-2,1,2)$ & $(63,10^{-1},10^{-2},10^{-3})$ & $(3,10,10,10^{-3})$ & $(13,10^{2},10^{-2},10^{-2},-1.5,1.5)$ \\\hline

Chimpanzee\_vs\_Leopard & 60.000 & 63.333 & 65.000 & 65.000 & 63.333 & 80.000 & 76.667 & \textbf{86.667} \\
$\big(200 \times(131\ \&\ 96)\big)$ & $(23,10^{-3})$ & $(33,10^{-3})$ & $(3,10^{-3})$ & $(43,10^{-3})$ & $(100,10^{-2},1,2,1,2)$ & $(63,10^{-3},10,10^{-3})$ & $(3,10,10,10^{-3})$ & $(3,10^{2},10^{-1},10^{-2},0.5,0.5)$ \\\hline

Chimpanzee\_vs\_Persiancat & 60.000 & 58.333 & 73.333 & 61.667 & 81.667 & 71.667 & \textbf{90.000} & 85.000 \\
$\big(200 \times(131\ \&\ 96)\big)$ & $(53,10^{-3})$ & $(13,1)$ & $(3,10^{-3})$ & $(33,10^{-3})$ & $(100,10^{-2},-2,1,1,2)$ & $(23,10,10^{-3},10^{-3})$ & $(3,1,1,10^{-3})$ & $(3,10,10^{-2},10^{-3},-1.5,0.5)$ \\\hline

Chimpanzee\_vs\_Pig & 71.667 & 75.000 & 63.333 & 66.667 & 70.000 & 75.000 & 76.667 & \textbf{78.333} \\
$\big(200 \times(131\ \&\ 96)\big)$ & $(33,1)$ & $(33,10^{-3})$ & $(13,10^{-3})$ & $(43,10^{-1})$ & $(1,10^{-2},-2,-2,2,2)$ & $(3,10^{2},10^{-1},10^{-3})$ & $(23,10^{-1},10^{-1},10^{-2})$ & $(3,10^{-3},10^{-2},10^{-3},-1.5,0.5)$ \\\hline

Chimpanzee\_vs\_Raccoon & 60.000 & 65.000 & 70.000 & 68.333 & 73.333 & 65.000 & \textbf{80.000} & \textbf{80.000} \\
$\big(200 \times(131\ \&\ 96)\big)$ & $(23,10^{-3})$ & $(3,10^{-3})$ & $(13,10^{-2})$ & $(3,10^{-3})$ & $(1,10^{-2},-2,2,2,2)$ & $(43,10^{-3},10^{-3},10)$ & $(3,1,1,10^{-3})$ & $(23,10,10^{-3},10^{-3},0.5,1.5)$ \\\hline

Chimpanzee\_vs\_rat & 55.000 & 58.333 & 53.333 & 65.000 & 66.667 & \textbf{81.667} & 73.333 & 71.667 \\
$\big(200 \times(131\ \&\ 96)\big)$ & $(63,10^{-1})$ & $(13,10^{-3})$ & $(3,10^{-2})$ & $(33,10^{-3})$ & $(1,10^{-2},-2,1,0,2)$ & $(13,10^{-2},10^{2},10^{-2})$ & $(23,10,10,10^{-3})$ & $(3,10^{-2},10^{-2},10^{-3},-0.5,2)$ \\\hline

Giantpanda\_vs\_Hippopotamus & 71.667 & 70.000 & 71.667 & 75.000 & 71.667 & 86.667 & \textbf{93.333} & 86.667 \\
$\big(200 \times(131\ \&\ 96)\big)$ & $(43,10^{-3})$ & $(3,10^{-3})$ & $(13,10^{-3})$ & $(53,10^{-3})$ & $(50,10^{-2},-2,2,0,2)$ & $(13,10^{-2},10^{-2},10^{-3})$ & $(3,1,1,10^{-3})$ & $(13,10,10^{-2},10^{-3},-1.5,1.5)$ \\\hline

Giantpanda\_vs\_Humpbackwhale & 85.000 & 88.333 & 91.667 & 88.333 & 93.333 & 91.667 & \textbf{95.000} & 93.333 \\
$\big(200 \times(131\ \&\ 96)\big)$ & $(33,10^{-3})$ & $(63,10^{-3})$ & $(13,10^{-3})$ & $(13,10^{-3})$ & $(1,10^{-2},-2,-2,-2,2)$ & $(23,10,10^{2},10^{-1})$ & $(23,10^{-1},10^{-3},10^{-3})$ & $(3,10,10^{-3},10^{-3},-1.5,0.5)$ \\\hline

Giantpanda\_vs\_Pig & 55.000 & 70.000 & 63.333 & 56.667 & 68.333 & 75.000 & \textbf{76.667} & 75.000 \\
$\big(200 \times(131\ \&\ 96)\big)$ & $(63,10^{-3})$ & $(13,10^{-3})$ & $(3,10^{-1})$ & $(63,1)$ & $(1,10^{-2},2,2,-1,2)$ & $(3,10^{-1},10^{-1},10^{-3})$ & $(3,1,1,10^{-1})$ & $(3,10,10^{-3},10^{-3},-1.5,1)$ \\\hline

Leopard\_vs\_Hippopotamus & 71.667 & 70.000 & 71.667 & 75.000 & 71.667 & 60.000 & \textbf{81.667} & 80.000 \\
$\big(200 \times(131\ \&\ 96)\big)$ & $(43,10^{-3})$ & $(3,10^{-3})$ & $(13,10^{-3})$ & $(53,10^{-3})$ & $(50,10^{-2},-2,2,0,2)$ & $(63,10,10^{-1},10^{-3})$ & $(43,1,10^{-3},10^{-3})$ & $(43,10,10^{2},10,0.5,1)$ \\\hline

Leopard\_vs\_Humpbackwhale & 85.000 & 88.333 & 91.667 & 88.333 & 93.333 & 93.333 & \textbf{100.000} & 96.667 \\
$\big(200 \times(131\ \&\ 96)\big)$ & $(33,10^{-3})$ & $(63,10^{-3})$ & $(13,10^{-3})$ & $(13,10^{-3})$ & $(1,10^{-2},-2,-2,-2,2)$ & $(13,10^{-3},10^{-3},10^{-3})$ & $(13,1,1,10^{-3})$ & $(3,10^{2},10^{-1},10^{-3},-1.5,0.5)$ \\\hline

Leopard\_vs\_Persiancat & 75.000 & 81.667 & 75.000 & 78.333 & 76.667 & 80.000 & \textbf{90.000} & \textbf{90.000} \\
$\big(200 \times(131\ \&\ 96)\big)$ & $(43,10^{-3})$ & $(33,10^{-3})$ & $(13,10^{-2})$ & $(53,10^{-3})$ & $(100,10^{-2},-2,2,1,2)$ & $(53,10^{3},10^{4},10)$ & $(3,1,1,10^{-3})$ & $(3,10,10^{-3},10^{-3},-0.5,0.5)$ \\\hline

Leopard\_vs\_Pig & 55.000 & 70.000 & 63.333 & 56.667 & 66.667 & 63.333 & \textbf{78.333} & \textbf{78.333} \\
$\big(200 \times(131\ \&\ 96)\big)$ & $(63,10^{-3})$ & $(13,10^{-3})$ & $(3,10^{-1})$ & $(63,1)$ & $(1,10^{-2},2,2,-1,2)$ & $(33,10^{2},10^{-2},10)$ & $(3,1,1,10^{-3})$ & $(3,10^{-2},10,10^{-3},-0.5,2)$ \\\hline

Persiancat\_vs\_Hippopotamus & 68.333 & 70.000 & 66.667 & 73.333 & 86.667 & 72.667 & 81.667 & \textbf{88.333} \\
$\big(200 \times(131\ \&\ 96)\big)$ & $(43,10^{-3})$ & $(33,10^{-2})$ & $(23,10^{-3})$ & $(43,10^{-2})$ & $(1,10^{-2},-2,-2,0,2)$ & $(13,10^{3},10^{3},10^{-4})$ & $(3,10,10,10^{-2})$ & $(13,10,10^{-3},10^{-3},-0.5,0.5)$ \\\hline

Persiancat\_vs\_Humpbackwhale & 73.333 & \textbf{95.000} & 78.333 & 81.667 & 83.333 & 81.000 & 93.333 & \textbf{95.000} \\
$\big(200 \times(131\ \&\ 96)\big)$ & $(23,10^{-3})$ & $(33,10^{-3})$ & $(3,10^{-3})$ & $(43,10^{-3})$ & $(100,10^{-2},1,2,2,2)$ & $(33,10^{-3},10^{-3},10^{-1})$ & $(13,1,1,10^{-3})$ & $(3,10,10^{-2},10^{-3},-1.5,0.5)$ \\\hline

Persiancat\_vs\_Pig & 60.000 & 66.667 & 55.000 & 58.333 & \textbf{75.000} & 58.333 & 68.333 & 63.333 \\
$\big(200 \times(131\ \&\ 96)\big)$ & $(53,10^{-3})$ & $(3,10^{-3})$ & $(3,10^{-3})$ & $(13,10^{-3})$ & $(50,10^{-2},2,-2,1,2)$ & $(43,10,10^{-1},10^{-2})$ & $(43,1,1,10^{-2})$ & $(3,10^{-1},10^{-1},10^{-3},-1.5,0.5)$ \\\hline

Pig\_vs\_Hippopotamus & 56.667 & 63.333 & 60.000 & 58.333 & 60.000 & 60.000 & 65.000 & \textbf{66.667} \\
$\big(200 \times(131\ \&\ 96)\big)$ & $(43,10^{-3})$ & $(43,10^{-3})$ & $(3,10^{-3})$ & $(53,10^{-3})$ & $(1,10^{-2},-2,-2,-1,2)$ & $(33,10^{-2},10^{-3},10^{-2})$ & $(43,1,1,10^{-3})$ & $(23,10^{-3},10^{-3},10^{-3},-0.5,0.5)$ \\\hline

Pig\_vs\_Humpback & 81.667 & 85.000 & 86.667 & \textbf{91.667} & 88.333 & 80.000 & 86.667 & 88.333 \\
$\big(200 \times(131\ \&\ 96)\big)$ & $(53,10^{-3})$ & $(3,10^{-3})$ & $(13,10^{-3})$ & $(13,10^{-3})$ & $(1,10^{-2},-2,-2,0,2)$ & $(33,10^{2},10^{-3},10)$ & $(23,1,1,10^{-3})$ & $(43,10^{-3},10^{2},10^{-3},-0.5,0.5)$ \\\hline

Raccoon\_vs\_Hippopotamus & 58.333 & 76.667 & \textbf{81.667} & 78.333 & 73.333 & \textbf{81.667} & 70.000 & 71.667 \\
$\big(200 \times(131\ \&\ 96)\big)$ & $(43,10^{-3})$ & $(13,10^{-3})$ & $(13,10^{-3})$ & $(3,10^{-3})$ & $(1,10^{-2},2,2,2,2)$ & $(13,10^{-3},10,10^{-3})$ & $(13,10^{-1},10^{-1},10^{-3})$ & $(53,10,10^{-2},10^{-2},-1.5,1.5)$ \\\hline

Raccoon\_vs\_Humpback & 85.000 & 86.667 & 86.667 & 85.000 & 93.333 & 88.333 & 95.000 & \textbf{96.667} \\
$\big(200 \times(131\ \&\ 96)\big)$ & $(23,10^{-3})$ & $(33,10^{-3})$ & $(13,10^{-3})$ & $(23,10^{-3})$ & $(1,10^{-2},-2,-2,-2,2)$ & $(43,10^{-2},10^{-2},10^{-2})$ & $(3,10^{-1},10^{-1},10^{-2})$ & $(13,10^{2},10^{-2},10^{-3},-1.5,1.5)$ \\\hline

Raccoon\_vs\_Leopard & 45.000 & 53.333 & 58.333 & 65.000 & 58.333 & 63.333 & \textbf{71.667} & 68.333 \\
$\big(200 \times(131\ \&\ 96)\big)$ & $(43,10^{-2})$ & $(33,10^{-3})$ & $(13,10^{-3})$ & $(43,10^{-3})$ & $(1,10^{-2},-2,-2,0,2)$ & $(53,10,10^{2},10)$ & $(3,1,1,10^{-3})$ & $(3,10^{-3},10^{-3},10^{-3},-1.5,0.5)$ \\\hline

Raccoon\_vs\_Rat & 56.667 & 60.000 & 56.667 & 55.000 & 65.000 & 59.330 & 73.333 & \textbf{85.000} \\
$\big(200 \times(131\ \&\ 96)\big)$ & $(63,10^{-3})$ & $(53,10^{-3})$ & $(63,10^{-3})$ & $(13,10^{-3})$ & $(50,10^{-2},-2,-2,1,2)$ & $(35,10^{-3},10^{-4},10)$ & $(53,1,1,10^{-3})$ & $(13,10,10^{-2},10^{-3},0.5,0.5)$ \\\hline

\textbf{Average AUC $\uparrow$} & 66.670 & 72.250 & 70.720 & 71.450 & 75.430 & 73.540 & 81.380 & \textbf{82.680} \\ \hline
\textbf{Average Rank $\downarrow$} & 7.02 & 5.13 & 5.70 & 5.13 & 4.02 & 4.78 & 2.24 & \textbf{1.98} \\
\hline
\multicolumn{9}{l}{Here, $^*$ denotes the proposed model.}\\
\end{tabular}
}
\end{adjustwidth}
\end{table*}
\end{landscape}

\noindent
With respect to
Giantpanda\_vs\_Pig, Leopard\_vs\_Hippopotamus, and
Leopard\_vs\_Humpbackwhale, the corresponding AUC values obtained by XGRVFL-MV are 75.000\%, 80.000\%, and 96.667\%, compared with 76.667\%, 81.667\%, and 100.000\% achieved by GRVFL-MV. 

MV-LDM records the highest AUC of 75.000\% on
Persiancat\_vs\_Pig, RVFL2 achieves 91.667\% on
Pig\_vs\_Humpback, RVFL1 and MvRVFL jointly obtain
81.667\% on Raccoon\_vs\_Hippopotamus, and
GRVFL-MV records 71.667\% on
Raccoon\_vs\_Leopard, while XGRVFL-MV obtains
63.333\%, 88.333\%, 71.667\%, and 68.333\%,
respectively. 
Overall, XGRVFL-MV records the highest average AUC together with the lowest average rank across all tasks, indicating consistent performance over the AwA benchmark datasets.

\subsection{Results on Corel5k Dataset}

The Corel5k dataset is derived from the Corel image collection and is commonly used for evaluating image classification and retrieval methods~\cite{datta2008image}. In this study, 5000 images from 50 classes are considered, with each class containing 100 images. The selected images represent diverse semantic categories, including
beaches, buildings, animals, and natural scenes. Visual features are extracted using MPEG-7 feature descriptors~\cite{eidenberger2004statistical},
where the 32-dimensional Color Structure Descriptor (CSD) is used as View~1 and the 64-dimensional Scalable Color Descriptor (SCD) is used as View~2.

% Following \cite{tanveer2025grvfl}, binary classification tasks are constructed by selecting one class as the positive class (100 samples) and randomly choosing 100 samples from the remaining classes as the negative class.

Table~\ref{table:resultscorel} reports the AUC (\%) obtained by the proposed XGRVFL-MV model and baseline methods, namely RVFLWoDL1, RVFLWoDL2, RVFL1, RVFL2, MV-LDM, MvRVFL, and GRVFL-MV, on 12 binary classification tasks constructed from the Corel5k dataset. Overall, XGRVFL-MV achieves the highest average AUC of 80.555\% and the lowest average rank of 2.33, followed by MV-LDM and GRVFL-MV with average AUC values of 79.999\% and 79.722\%, respectively, and an
average rank of 2.95. XGRVFL-MV achieves the highest AUC with respect to 100000 (75.000\%) and shares the highest AUC with GRVFL-MV With respect to 17000 (81.667\%), with RVFLWoDL2 and RVFL2 on task 122000 (81.667\%), and with MV-LDM on task 174000 (78.333\%). On the remaining tasks, the proposed model generally produces AUC values close to those of the best-performing baseline. With respect to 10000, XGRVFL-MV records 83.333\% compared with 86.667\% achieved by GRVFL-MV, while on task 13000 it achieves 85.000\% compared with 88.333\% obtained by MV-LDM. Similarly, on tasks 103000 and 108000, MV-LDM records the highest AUC values of 83.333\% and 91.667\%, whereas XGRVFL-MV obtains 81.667\% and 88.333\%, respectively.  On task 121000, RVFLWoDL2 and GRVFL-MV jointly achieve the highest AUC of 75.000\%, while XGRVFL-MV records 71.667\%, equal to the values obtained by MV-LDM and RVFL2. On task 130000, MV-LDM achieves the highest AUC of 88.333\%, whereas XGRVFL-MV and RVFLWoDL2 both obtain 85.000\%. With respect to 140000, GRVFL-MV records the highest AUC of 83.333\%, compared with 78.333\% obtained by XGRVFL-MV. With respect to 144000, RVFLWoDL1 achieves the highest AUC of 78.333\%, while XGRVFL-MV and GRVFL-MV both record 76.667\%. Overall,  XGRVFL-MV attains the highest average AUC together with the lowest average rank across all
Corel5k classification tasks.

\begin{table*}[!t]
\caption{Performance comparison of baseline and the proposed XGRVFL-MV model on  Corel5k datasets}
\centering
\label{table:resultscorel}
\begin{adjustwidth}{-3.0cm}{4.5cm}
\setlength{\tabcolsep}{4pt} 
\renewcommand{\arraystretch}{1.6}
\resizebox{1.6\linewidth}{!}{%
\begin{tabular}{|c|c|c|c|c|c|c|c|c|}
\hline
  & \textbf{RVFLWoDL1~\cite{huang2006extreme}} & \textbf{RVFLWoDL2~\cite{huang2006extreme}} & \textbf{RVFL1~\cite{pao1994learning}} & \textbf{RVFL2~\cite{pao1994learning}} & \textbf{MV-LDM~\cite{hu2024multiview}}& \textbf{MVRVFL~\cite{quadir2024multiview}} & \textbf{GRVFL-MV~\cite{tanveer2025grvfl}} & \textbf{XGRVFL-MV$^*$} \\\hline
\textbf{Dataset} & \textbf{AUC $\uparrow$} & \textbf{AUC $\uparrow$} & \textbf{AUC $\uparrow$} & \textbf{AUC $\uparrow$} & \textbf{AUC $\uparrow$} & \textbf{AUC $\uparrow$} & \textbf{AUC $\uparrow$} & \textbf{AUC $\uparrow$} \\
 $(Patterns \times Features)$ &  $(n, e)$ & $(n, e)$ & $(n, e)$ & $(n, e)$  & $(e, \epsilon,\lambda, v1, v2, \theta)$ & $(n, e1, e3, \rho)$ & $(n, e, \mu, \gamma)$ & $(n, e, \rho, \theta, a, \eta)$ \\\hline

10000 & 76.667 & 80.000 & 76.667 & 71.667 & 85.000 & 78.333 & \textbf{86.667} & 83.333 \\
$\big(200 \times(64\ \&\ 32)\big)$ & $(23,10^{-3})$ & $(53,10^{-2})$ & $(13,10^{-2})$ & $(43,10^{-3})$ & $(50,10^{-2},0,-2,0,1)$ & $(33,10^{-1},10^{-3},10^{-1})$ & $(33,10^{-2},10^{-3},10^{-3})$ & $(23,10^{-1},10^{-1},10^{-3},0.5,1.5)$ \\\hline

13000 & 80.000 & 78.333 & 83.333 & 76.667 & \textbf{88.333} & 73.333 & 83.333 & 85.000 \\
$\big(200 \times(64\ \&\ 32)\big)$ & $(53,10^{-3})$ & $(43,10^{-3})$ & $(53,10^{-2})$ & $(13,10^{-2})$ & $(10,10^{-2},-2,-2,0,1)$ & $(23,10^{-3},10^{-3},10^{-3})$ & $(33,10^{-1},10^{-3},10^{-3})$ & $(13,10^{-3},10^{-3},10^{-3},0.5,1.5)$ \\\hline

17000 & 76.667 & 75.000 & 73.333 & 73.333 & 68.333 & 73.333 & \textbf{81.667} & \textbf{81.667} \\
$\big(200 \times(64\ \&\ 32)\big)$ & $(63,10^{-3})$ & $(53,10^{-2})$ & $(63,10^{-3})$ & $(63,10^{-3})$ & $(1,10^{-2},-2,-2,-1,0)$ & $(43,10^{-3},10^{-3},10^{-3})$ & $(3,10^{-1},10^{-3},10^{-3})$ & $(43,10^{-1},10^{-2},10^{-3},0.5,1)$ \\\hline

100000 & 70.000 & 61.667 & 71.667 & 66.667 & 71.667 & 58.333 & 71.667 & \textbf{75.000} \\
$\big(200 \times(64\ \&\ 32)\big)$ & $(43,10^{-1})$ & $(23,10^{-3})$ & $(3,10^{-3})$ & $(3,10^{-3})$ & $(1,10^{-2},-2,1,-1,0)$ & $(95,10^{-3},10^{-3},10^{-3})$ & $(3,10^{2},10^{-3},10^{-3})$ & $(63,10^{-1},10^{-3},10^{-3},-0.5,1)$ \\\hline

103000 & 71.667 & 70.000 & 73.333 & 75.000 & \textbf{83.333} & 78.333 & 80.000 & 81.667 \\
$\big(200 \times(64\ \&\ 32)\big)$ & $(43,10^{-3})$ & $(23,10^{-2})$ & $(13,10^{-2})$ & $(23,10^{-2})$ & $(1,10^{-2},2,2,-1,0)$ & $(43,10^{-3},10^{-3},10^{-3})$ & $(33,10^{-1},10^{-3},10^{-2})$ & $(33,10^{-3},10^{-1},10^{-3},1,1.5)$ \\\hline

108000 & 81.667 & 78.333 & 83.333 & 80.000 & \textbf{91.667} & 81.667 & 86.667 & 88.333 \\
$\big(200 \times(64\ \&\ 32)\big)$ & $(53,10^{-3})$ & $(53,10^{-3})$ & $(3,10^{-3})$ & $(23,10^{-3})$ & $(1,10^{-2},2,1,0,-1)$ & $(43,10^{-3},10^{-3},10^{-3})$ & $(33,10^{-2},10^{-3},10^{-3})$ & $(23,10^{2},10^{-1},10^{-3},-0.5,1)$ \\\hline

121000 & 63.333 & \textbf{75.000} & 71.667 & 71.667 & 71.667 & 68.333 & \textbf{75.000} & 71.667 \\
$\big(200 \times(64\ \&\ 32)\big)$ & $(53,10^{-3})$ & $(63,10^{-3})$ & $(3,10^{-2})$ & $(13,10^{-3})$ & $(10,10^{-2},1,-2,0,2)$ & $(63,10^{-3},10^{-3},10^{-3})$ & $(33,10^{-2},10^{-3},10^{-3})$ & $(23,10^{-1},10^{-1},10^{-3},0.5,0.5)$ \\\hline

122000 & 78.333 & \textbf{81.667} & 80.000 & \textbf{81.667} & 80.000 & 63.333 & 75.000 & \textbf{81.667} \\
$\big(200 \times(64\ \&\ 32)\big)$ & $(53,10^{-3})$ & $(33,10^{-2})$ & $(3,10^{-3})$ & $(33,10^{-3})$ & $(1,10^{-2},-2,2,-1,0)$ & $(63,10^{-3},10^{-3},10^{-3})$ & $(13,10^{-1},10^{-3},10^{-2})$ & $(23,10^{-3},10^{-3},10^{-3},0.5,1)$ \\\hline

130000 & 73.333 & 80.000 & 75.000 & 78.333 & \textbf{88.333} & 73.333 & 85.000 & 85.000 \\
$\big(200 \times(64\ \&\ 32)\big)$ & $(53,10^{-3})$ & $(33,10^{-3})$ & $(63,10^{-3})$ & $(53,10^{-2})$ & $(1,10^{-2},1,-2,0,0)$ & $(33,10^{-3},10^{-3},10^{-3})$ & $(3,10^{-2},10^{-3},10^{-3})$ & $(3,10^{-1},10^{-3},10^{-3},1,1.5)$ \\\hline

140000 & 73.333 & 78.333 & 78.333 & 71.667 & 78.333 & 76.667 & \textbf{83.333} & 78.333 \\
$\big(200 \times(64\ \&\ 32)\big)$ & $(63,10^{-3})$ & $(63,10^{-2})$ & $(3,1)$ & $(63,10^{-3})$ & $(10,10^{-2},-2,-2,-2,1)$ & $(53,10^{-3},10^{-3},10^{-3})$ & $(3,10^{-1},10^{-3},10^{-2})$ & $(43,10^{-3},10^{-3},10^{-3},-1,0.5)$ \\\hline

144000 & \textbf{78.333} & 75.000 & 73.333 & 73.333 & 75.000 & 73.333 & 76.667 & 76.667 \\
$\big(200 \times(64\ \&\ 32)\big)$ & $(63,10^{-3})$ & $(23,10^{-1})$ & $(3,1)$ & $(13,10^{-3})$ & $(50,10^{-2},0,2,-1,2)$ & $(43,10^{-3},10^{-3},10^{-3})$ & $(53,10^{-2},10^{-3},10^{-3})$ & $(3,10^{-3},10^{-1},10^{-3},0.5,0.5)$ \\\hline

174000 & 63.333 & 76.667 & 53.333 & 73.333 &  \textbf{78.333} & 73.333 & 71.667 & \textbf{78.333} \\
$\big(200 \times(64\ \&\ 32)\big)$ & $(43,10^{-2})$ & $(63,10^{-3})$ & $(23,1)$ & $(3,10^{-3})$ & $(10,10^{-2},0,-2,-2,1)$ & $(23,10^{-3},10^{-3},10^{-3})$ & $(13,10^{-1},10^{-2},10^{-2})$ & $(23,10^{-3},10^{-1},10^{-3},0.5,1)$ \\\hline

\textbf{Average AUC $\uparrow$} & 73.888 & 75.833 & 74.444 & 74.444 & 79.999 & 72.638 & 79.722 & \textbf{80.555} \\ \hline
\textbf{Average Rank $\downarrow$} & 5.70 & 4.62 & 5.20 & 5.83 & 2.95 & 6.37 & 2.95 & \textbf{2.33} \\
\hline
\multicolumn{9}{l}{Here, $^*$ denotes the proposed model.}\\
\end{tabular}
}
\end{adjustwidth}
\end{table*}

\subsection{Statistical Analysis}

Statistical analyses are performed using the results obtained on the UCI and KEEL datasets. The Friedman test, the Wilcoxon signed-rank test, and the win-tie-loss comparison are used to compare the
performance of the proposed XGRVFL-MV model with the baseline methods.

\subsubsection{Friedman Analysis}

We apply the Friedman test~\cite{friedman1940comparison} to compare the
performance of the baseline methods on the UCI and KEEL benchmark datasets. The Friedman test is a non-parametric statistical procedure that compares multiple algorithms using their average ranks across different datasets. Since the test operates on rankings rather than the original performance values, it does not require the assumption of data normality. Table~\ref{table:results1} reports the AUC values obtained by the compared methods. Based on these results, the average ranks of RVFLWoDL1, RVFLWoDL2, RVFL1, RVFL2, MV-LDM, MvRVFL, GRVFL-MV, and XGRVFL-MV are 5.28, 5.03, 5.78, 4.69, 3.42, 6.17, 3.42, and 2.22, respectively. We compute the Friedman chi-square statistic as

\begin{equation}
\chi^2_F =
\frac{12N}{K(K+1)}
\left[
\sum_{j=1}^{K}R_j^2
-
\frac{K(K+1)^2}{4}
\right],
\end{equation}

where $N=18$ denotes the number of datasets and $K=8$ denotes the
number of compared methods. The computed Friedman statistic is
$\chi^2_F = 38.9217$.
To obtain a more accurate approximation for a finite number of datasets,
we transform the Friedman statistic into the Iman-Davenport
F-statistic,

\begin{equation}
F_F =
\frac{(N-1)\chi^2_F}
{N(K-1)-\chi^2_F}.
\end{equation}

\begin{figure}[H]
  \centering
\includegraphics[width=0.9\textwidth]{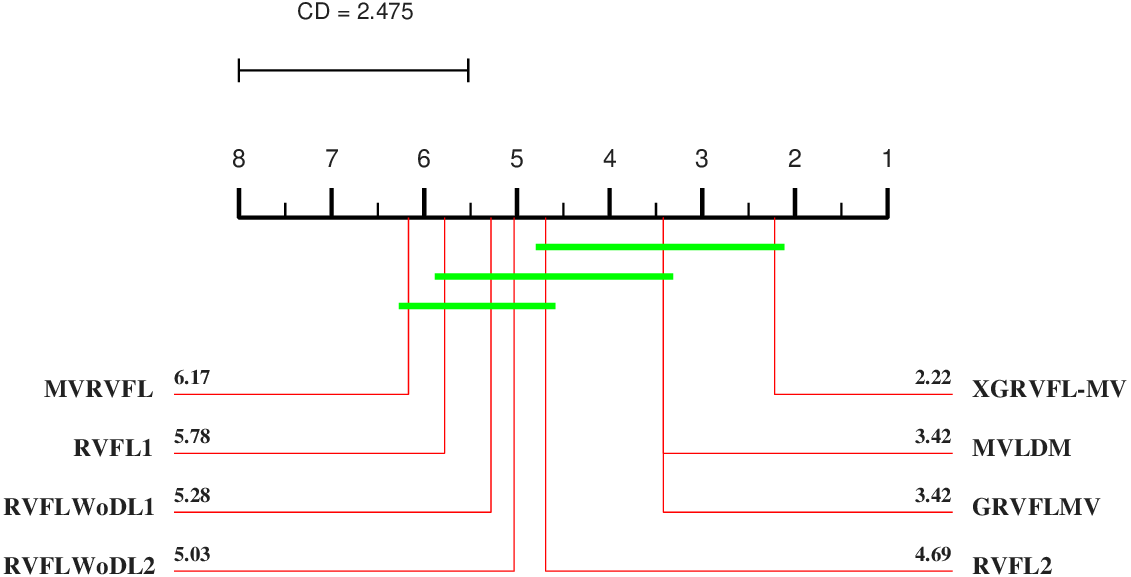}
  \caption{Average ranks of the compared methods on the UCI and KEEL datasets illustrated using a critical difference (CD) diagram based on the Nemenyi post-hoc test.}
  \label{fig:cd}
\end{figure}

\noindent
The computed value is $F_F=7.5986$. At the 5\% significance level, this
value exceeds the corresponding critical value of the
$F(7,119)$ distribution. Therefore, we reject the null hypothesis that
all compared methods perform equivalently. This result indicates a
statistically significant difference among the compared methods. Figure~\ref{fig:cd} presents the critical difference (CD) diagram
obtained from the Nemenyi post-hoc test using the average ranks of the
compared methods. We compute the critical difference as

\begin{equation}
CD =
q_{\alpha}
\sqrt{\frac{K(K+1)}{6N}},
\end{equation}

where $q_{\alpha=0.05}=3.031$ for $K=8$ methods. The resulting critical difference is $CD=2.4700$. The Nemenyi post-hoc test identifies pairs of methods whose average rank
differences exceed the computed critical difference. Based on the CD
diagram, XGRVFL-MV differs significantly from RVFLWoDL1, RVFLWoDL2,
RVFL1, RVFL2, and MvRVFL. In contrast, the differences between
XGRVFL-MV and MV-LDM, and between XGRVFL-MV and GRVFL-MV, do not exceed
the critical difference. Among all compared methods, XGRVFL-MV attains
the lowest average rank (2.22). It also achieves the highest average
AUC (79.530\%) reported in Table~\ref{table:results1}.

\subsubsection{Wilcoxon Analysis}

We use the Wilcoxon signed-rank test~\cite{demvsar2006statistical} to
perform pairwise comparisons between XGRVFL-MV and each baseline method over the UCI and KEEL benchmark datasets. The Wilcoxon signed-rank test is a non-parametric statistical procedure that compares two methods using the ranked differences in their paired performance values. For each comparison, the test computes the positive rank sum ($R^{+}$) and the negative rank sum ($R^{-}$), defined as:

\begin{equation}
R^{+}=
\sum_{d_i>0}\mathrm{rank}(d_i)
+\frac{1}{2}\sum_{d_i=0}\mathrm{rank}(d_i),
\end{equation}

\begin{equation}
R^{-}=
\sum_{d_i<0}\mathrm{rank}(d_i)
+\frac{1}{2}\sum_{d_i=0}\mathrm{rank}(d_i),
\end{equation}
\noindent
where $d_i$ denotes the performance difference between XGRVFL-MV and the baseline method on the $i^{\mathrm{th}}$ dataset. A $p$-value smaller than 0.05 indicates a statistically significant difference between the two compared methods. 

\begin{table}[htbp]
\caption{Wilcoxon signed-rank test results (\(R^{+}\), \(R^{-}\), and \(p\)-values) for XGRVFL-MV and the baseline methods.}
    \begin{center}
    \begin{tabular}{|c|c|c|c|c|c|}
        \hline 
       Model & $R+$ & $R-$ & $p$-value & Hypothesis (0.05) \\
        \hline    
        RVFLWoDL1 \cite{huang2006extreme}  & 162.50 & 8.50 & 0.000237 & R \\
         \hline  
        RVFLWoDL2 \cite{huang2006extreme}  & 158.00 & 13.00 & 0.000641 & R \\
         \hline  
        RVFL1 \cite{pao1994learning} & 171.00 & 0.00 & 0.000008 & R \\
         \hline  
        RVFL2 \cite{pao1994learning}  & 159.00 & 12.00 & 0.000526 & R \\
         \hline  
        MV-LDM \cite{hu2024multiview}  & 119.00 & 52.00 & 0.147949 & {NR} \\
         \hline 
       MVRVFL \cite{quadir2024multiview} & 160.00 & 11.00 & 0.000404 &  R \\
        \hline
        GRVFL-MV \cite{tanveer2025grvfl} & 123.00 & 48.00 & 0.106071 &  NR \\
        \hline
     \multicolumn{5}{|l|}{\textbf{R:} Rejected, \textbf{NR:} Not Rejected (Null Hypothesis)}\\
        \hline
    \end{tabular}
 \label{tab:my_label1}
 \end{center}
\end{table}

For every pairwise comparison,
the positive rank sum ($R^{+}$) is greater than the corresponding negative rank sum ($R^{-}$). The values of $R^{+}$ range from 119.00 to 171.00, whereas the values of $R^{-}$ range from 0.00 to 52.00. The comparisons with RVFLWoDL1 ($p=0.000237$), RVFLWoDL2
($p=0.000641$), RVFL1 ($p=0.000080$), RVFL2 ($p=0.000526$), and MvRVFL ($p=0.000404$) produce $p$-values below the 0.05 significance level. Therefore, we reject the null hypothesis for these pairwise
comparisons, indicating statistically significant differences between
XGRVFL-MV and these methods.

\subsubsection{Pairwise Win-Tie-Loss Analysis}

We perform a pairwise Win-Tie-Loss analysis~\cite{demvsar2006statistical} to compare XGRVFL-MV with the baseline methods across the UCI and KEEL
datasets. For each pairwise comparison, the analysis counts the number of datasets on which one method achieves a higher performance value (win), the
same performance value (tie), or a lower performance value (loss). This analysis provides a dataset-wise comparison that complements the Friedman
and Wilcoxon statistical tests. We compare XGRVFL-MV with baseline methods: RVFLWoDL1, RVFLWoDL2,
RVFL1, RVFL2, MV-LDM, MvRVFL, and GRVFL-MV. Table~\ref{win-lie-loss} summarizes the Win-Tie-Loss counts for all pairwise comparisons over the
18 datasets.

\begin{table}[htbp]
\caption{Shows the results of the Win-Tie-Loss test applied to baseline models and the proposed model.}
\centering
\begin{adjustwidth}{-1.5cm}{-1.5cm}    % increase if still overflowing
\setlength{\tabcolsep}{2pt} 
\renewcommand{\arraystretch}{1.25} 
\resizebox{\linewidth}{!}{%
\begin{tabular}{|c|c|c|c|c|c|c|c|}
\hline
Models & RVFLWoDL1 & RVFLWoDL2 & RVFL1 & RVFL2 &  MV-LDM & MVRVFL & GRVFL-MV \\
\hline
RVFLWoDL2 \cite{huang2006extreme} & $[9,1,8] $ & & & & & & \\
\hline
RVFL1 \cite{pao1994learning} & $[3,8,7]$ & $[7,1,10]$ & & & & &\\ 
\hline
RVFL2 \cite{pao1994learning} & $[10,1,7]$ & $[7,7,4]$ & $[10,3,5]$ & & & & \\
\hline
MV-LDM \cite{hu2024multiview} & $[14,0,4]$ & $[12,0,6]$ & $[14,0,4]$ &$[12,0,6]$ & & & \\
\hline
MVRVFL \cite{quadir2024multiview} & $[6,0,12]$ & $[5,0,13]$ & $[7,0,11]$ &$[4,0,14]$ & $[4,0,14]$ & & \\
\hline
GRVFL-MV \cite{tanveer2025grvfl} & $[14,0,4]$ & $[13,0,5]$ & $[14,0,4]$ & $[12,0,6]$ & $[7,1,10]$& $[14,0,4]$ &\\
\hline
XGRVFL-MV$^*$ & $[16,0,2]$ & $[16,0,2]$ & $[18,0,0]$ & $[17,0,1]$ & $[12,0,6]$ &$[15,0,3]$ & $[10,0,8]$ \\
\hline
 \multicolumn{8}{l}{Here, $^*$ denotes the proposed model.}
\end{tabular}
}
\end{adjustwidth}
\label{win-lie-loss}
\end{table}

The results show that XGRVFL-MV records more wins than losses against all baseline methods. Compared with RVFLWoDL1 and RVFLWoDL2, XGRVFL-MV records 16 wins and 2 losses, with no ties in either comparison. Against RVFL1 and RVFL2, it records 18-0-0 and 17-0-1 Win-Tie-Loss counts, respectively. The corresponding Win-Tie-Loss counts against MV-LDM, MvRVFL, and GRVFL-MV are 12-0-6, 15-0-3, and 10-0-8, respectively. For the pairwise Win-Tie-Loss analysis, the significance of the observed
win counts is assessed using the normal approximation of the sign test.
For \(N\) datasets, the corresponding significance threshold at the
5\% significance level is computed as
$
W_{\mathrm{crit}}
=
\frac{N}{2}
+
\frac{1.96\sqrt{N}}{2},
$
where \(N\) denotes the number of datasets. For the benchmark datasets, the threshold is
\(W_{\mathrm{crit}}=13.1578\). Accordingly, the observed win counts
against RVFLWoDL1, RVFLWoDL2, RVFL1, RVFL2, and MvRVFL exceed this
threshold, whereas the corresponding win counts against MV-LDM and
GRVFL-MV do not.

\subsection{Sensitivity Analysis of Hyperparameters on the UCI and KEEL Datasets}

\begin{figure}[H]
  \centering
  
  % First row
  \begin{minipage}[b]{0.47\textwidth}
    \centering
    \includegraphics[width=0.8\textwidth]{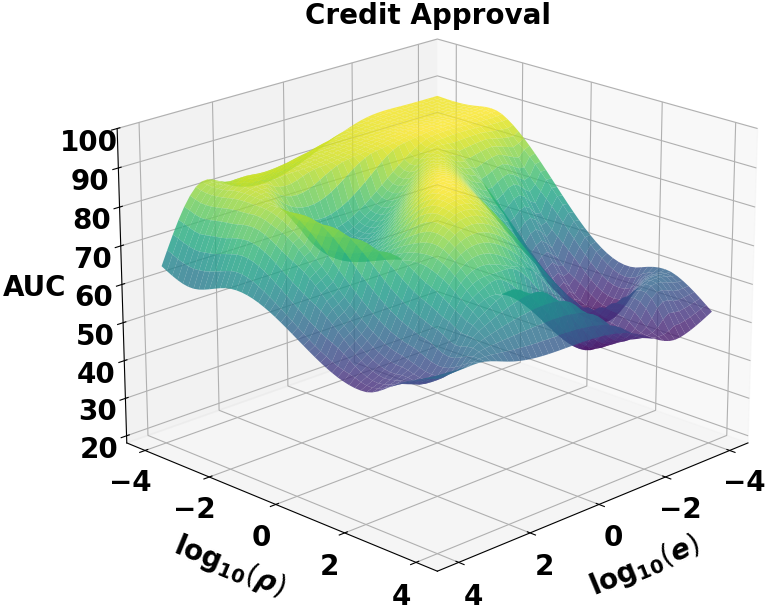}
  \end{minipage} \hfill
  \begin{minipage}[b]{0.47\textwidth}
    \centering
    \includegraphics[width=0.8\textwidth]{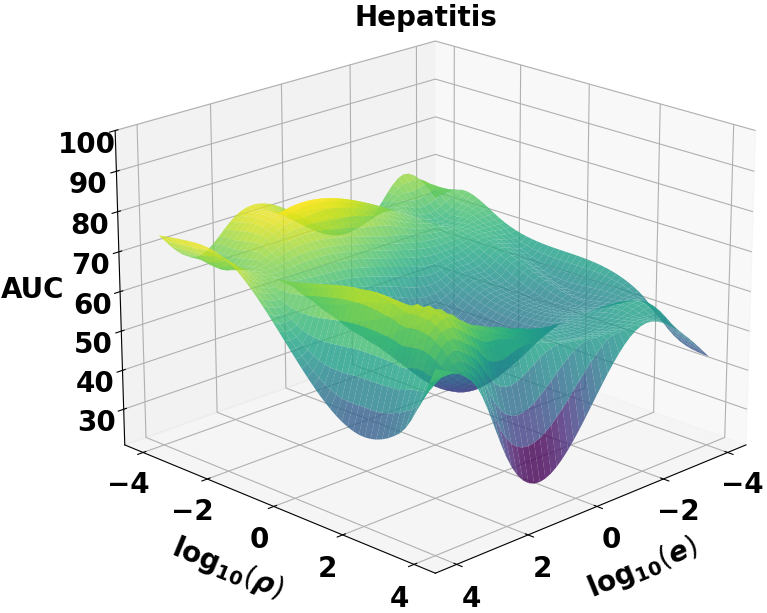}
  \end{minipage}\\[0.80ex]
  
  % Second row
  \begin{minipage}[b]{0.47\textwidth}
    \centering
    \includegraphics[width=0.8\textwidth]{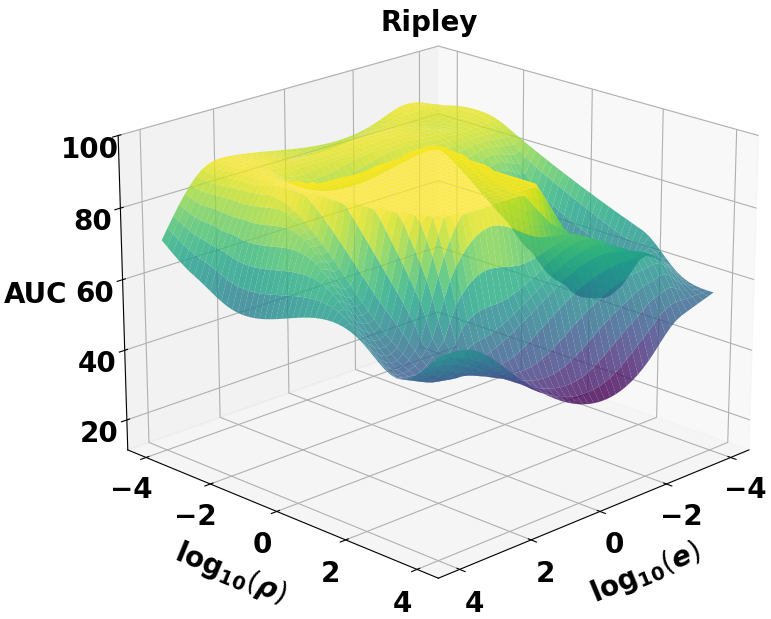}
  \end{minipage} \hfill
  \begin{minipage}[b]{0.47\textwidth}
    \centering
    \includegraphics[width=0.8\textwidth]{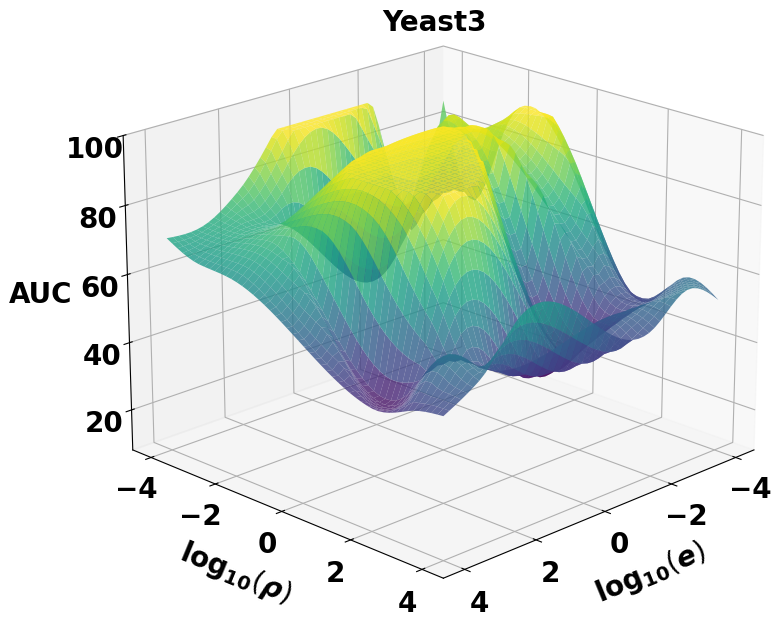}
  \end{minipage}

  \caption{AUC variation with $e$ and $\rho$ for the proposed model on UCI and KEEL datasets}
  \label{sensitivity_UCI}
\end{figure}

We analyze the sensitivity of XGRVFL-MV to the regularization parameters $\rho$ and $e$ on the UCI and KEEL datasets. For this analysis, parameters vary from $10^{-4}$ to $10^{4}$ while fixing the remaining hyperparameters at their selected values. Figure~\ref{sensitivity_UCI} shows the resulting AUC surfaces for four representative datasets: Credit Approval, Hepatitis, Ripley, and Yeast3. The horizontal axes represent $\log_{10}(\rho)$ and $\log_{10}(e)$, and the vertical axis reports the corresponding AUC values. For the Credit Approval dataset, the highest AUC values are observed over a broad region corresponding to small and moderate values of $\rho$ and $e$. As the value of $e$ increases, the AUC gradually decreases for many parameter combinations. For the Hepatitis dataset, the high-performance region occupies a comparatively smaller portion of the parameter space, and the AUC decreases for several combinations of larger $\rho$ and $e$.
The Ripley dataset exhibits relatively small performance variations over a wide range of parameter settings, with AUC values remaining within a comparatively narrow interval. In contrast, the Yeast3 dataset shows a more irregular response surface, where several parameter combinations produce similar high AUC values. Overall, the sensitivity analysis shows that the influence of $\rho$ and $e$ varies across datasets. Credit Approval and Ripley exhibit relatively stable performance over a wider range of parameter values, whereas Hepatitis and Yeast3 show larger variations in AUC across the parameter space. These observations indicate that selecting $\rho$ and $e$ through dataset-specific hyperparameter tuning is appropriate for the proposed
XGRVFL-MV model.

\subsection{Sensitivity Analysis of Hyperparameters on the AwA Datasets}

\begin{figure}[H]
  \centering
  
  % First row
  \begin{minipage}[b]{0.47\textwidth}
    \centering
    \includegraphics[width=0.8\textwidth]{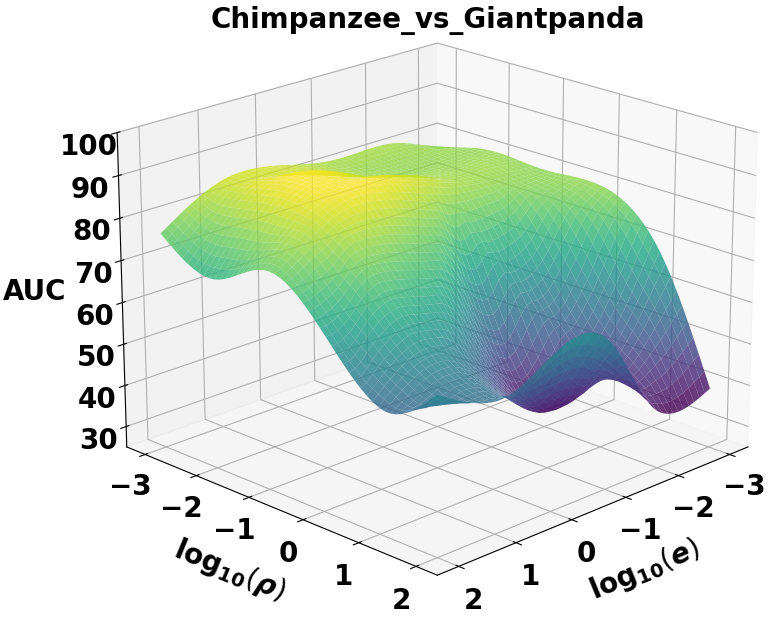}
  \end{minipage} \hfill
  \begin{minipage}[b]{0.47\textwidth}
    \centering
    \includegraphics[width=0.8\textwidth]{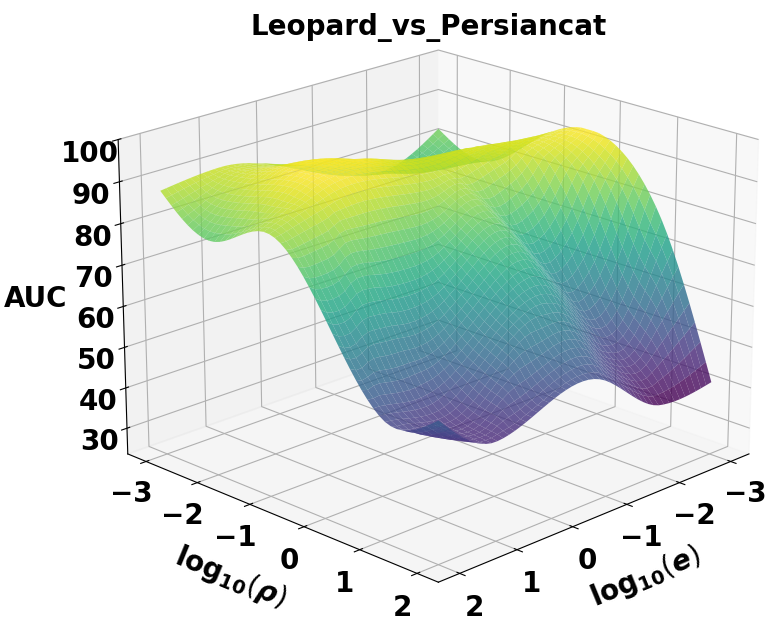}
  \end{minipage}\\[0.80ex]
  
  % Second row
  \begin{minipage}[b]{0.47\textwidth}
    \centering
    \includegraphics[width=0.8\textwidth]{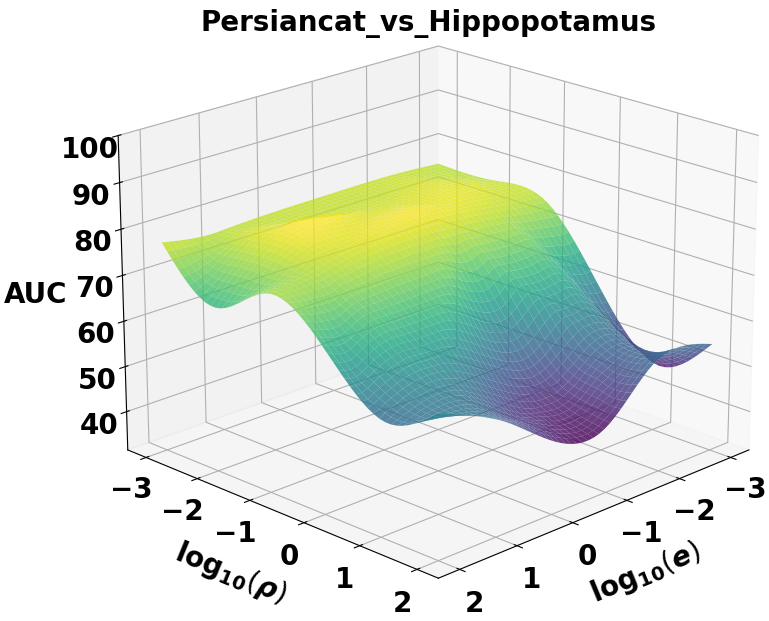}
  \end{minipage} \hfill
  \begin{minipage}[b]{0.47\textwidth}
    \centering
    \includegraphics[width=0.8\textwidth]{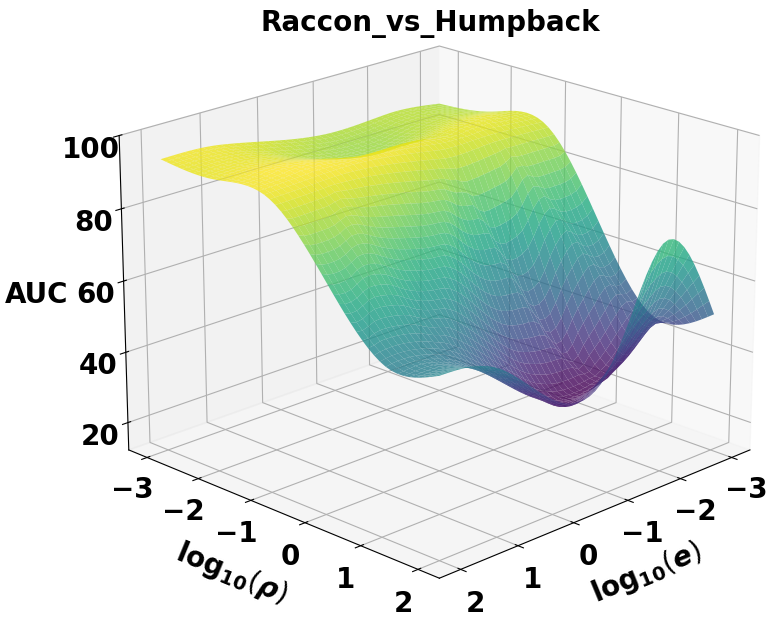}
  \end{minipage}

  \caption{AUC variation with $e$ and $\rho$ for the proposed model on Awa datasets}
  \label{sensitivity_awa}
\end{figure}

We analyze the sensitivity of XGRVFL-MV to the regularization parameters $\rho$ and $e$ on the AwA datasets. For this analysis, parameters vary from $10^{-3}$ to $10^{3}$ while fixing the remaining hyperparameters at their selected values. Figure~\ref{sensitivity_awa} shows the resulting AUC surfaces for four representative classification tasks: Chimpanzee\_vs\_Giantpanda, Leopard\_vs\_Persiancat, Persiancat\_vs\_Hippopotamus, and Raccoon\_vs\_Humpback. The horizontal axes correspond to $\log_{10}(\rho)$ and $\log_{10}(e)$, whereas the vertical axis reports
the AUC.

For Chimpanzee\_vs\_Giantpanda, high AUC values are observed over a broad region corresponding to small and moderate values of $\rho$ and $e$. The AUC decreases as the value of $e$ increases for several parameter combinations. A similar trend is observed for Leopard\_vs\_Persiancat, where high AUC values are obtained over a wide range of small and moderate parameter values, followed by lower AUC values as $e$ increases. The response surface for Persiancat\_vs\_Hippopotamus changes more gradually across the parameter space, with AUC values remaining above 60 for most parameter combinations. In contrast, Raccoon\_vs\_Humpback exhibits larger variations in AUC, with high values observed for several combinations of small $\rho$ and $e$ and lower AUC values appearing for larger parameter settings. Overall, the response surfaces show that the influence of $\rho$ and $e$ varies across the AwA classification tasks. Several tasks maintain high AUC values over a relatively broad range of parameter settings, whereas
others exhibit larger performance variations. These results indicate that selecting $\rho$ and $e$ through task-specific hyperparameter tuning is appropriate for the proposed XGRVFL-MV model.

\subsection{Sensitivity Analysis of Hyperparameters on the Corel5k Datasets}

\begin{figure}[H]
  \centering
  
  % First row
  \begin{minipage}[b]{0.47\textwidth}
    \centering
    \includegraphics[width=0.8\textwidth]{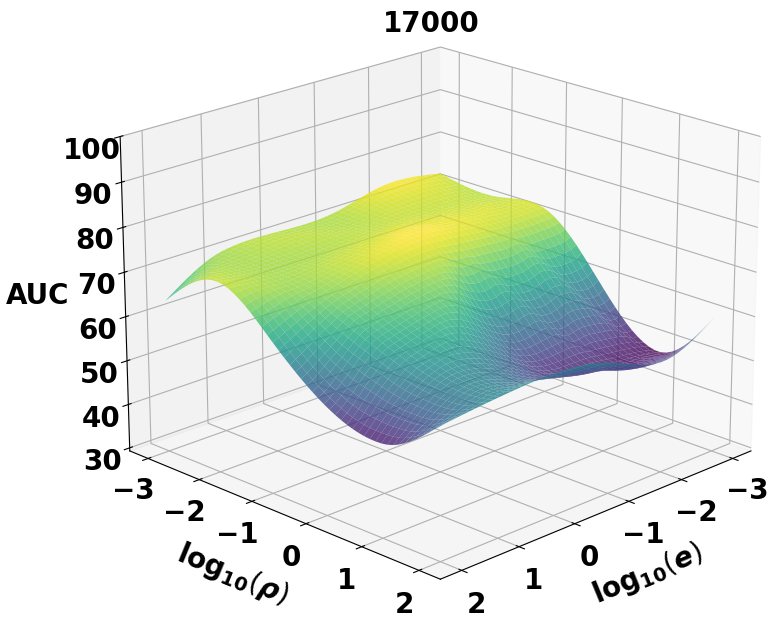}
  \end{minipage} \hfill
  \begin{minipage}[b]{0.47\textwidth}
    \centering
    \includegraphics[width=0.8\textwidth]{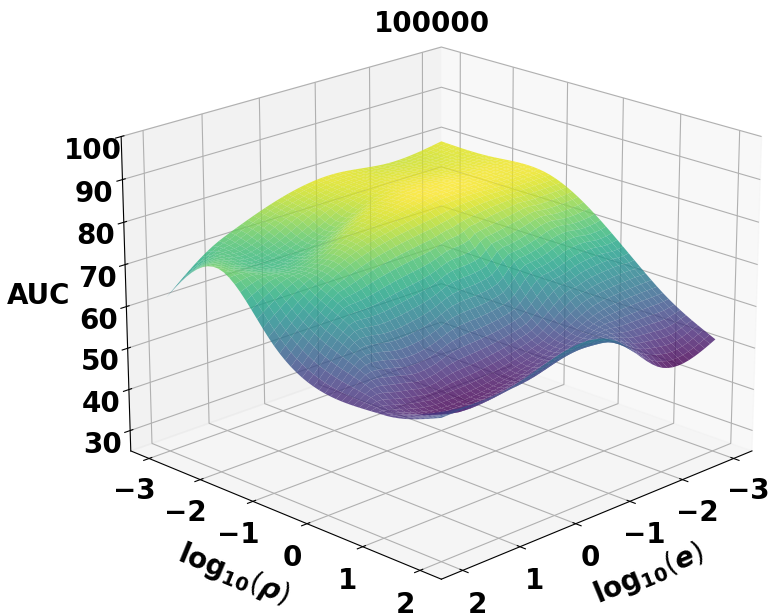}
  \end{minipage}\\[0.80ex]
  
  % Second row
  \begin{minipage}[b]{0.47\textwidth}
    \centering
    \includegraphics[width=0.8\textwidth]{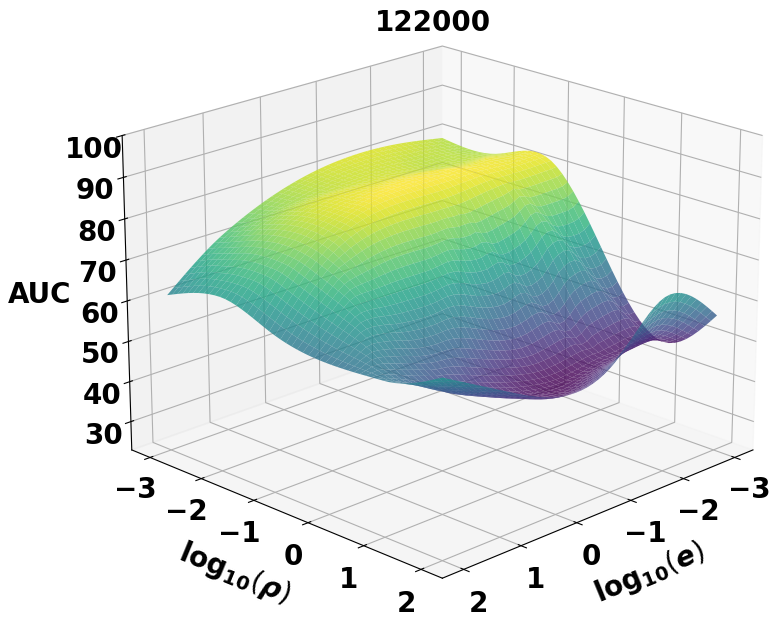}
  \end{minipage} \hfill
  \begin{minipage}[b]{0.47\textwidth}
    \centering
    \includegraphics[width=0.8\textwidth]{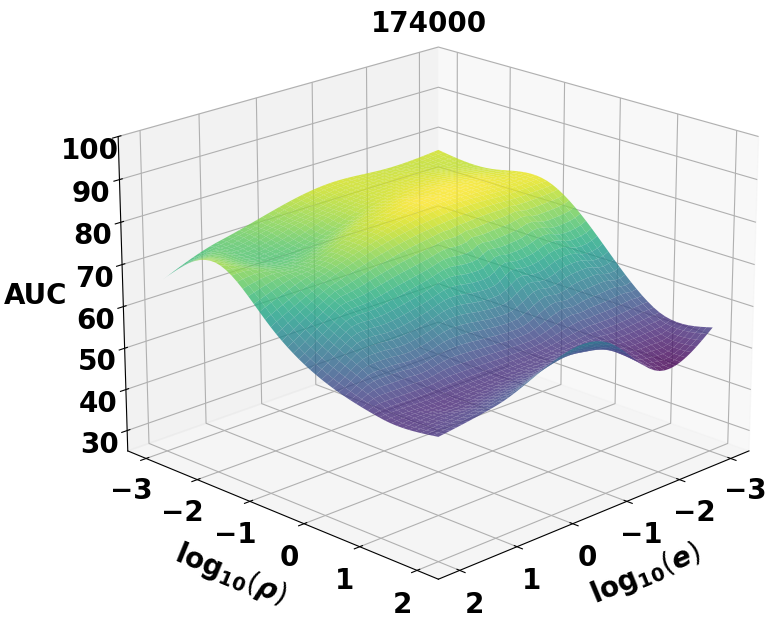}
  \end{minipage}

  \caption{AUC variation with $e$ and $\rho$ for the proposed model on Corel5k datasets}
  \label{sensitivity_Corel}
\end{figure}

We analyze the sensitivity of XGRVFL-MV to the regularization parameters $\rho$ and $e$ on the Corel5k datasets. For this analysis, we vary both parameters from $10^{-3}$ to $10^{2}$ while fixing the remaining hyperparameters at their selected values. Figure~\ref{sensitivity_Corel} shows the resulting AUC surfaces for four representative classification
tasks: 17000, 100000, 122000, and 174000. The horizontal axes represent $\log_{10}(\rho)$ and $\log_{10}(e)$, whereas the vertical axis reports the corresponding AUC values. For task 17000, high AUC values are observed over a broad region
corresponding to small and moderate values of $\rho$ and $e$. As the value of $e$ increases, the AUC gradually decreases for several parameter combinations. Task 100000 exhibits a comparatively smoother response surface, with AUC values remaining within a relatively narrow range across most of the parameter space and the highest values observed at small $\rho$ and $e$.  For task 122000, the high-AUC region is more localized, with AUC values above 90 occurring for several combinations of small $\rho$ and $e$, followed by lower AUC values as the parameter values increase.

Task 174000 exhibits a response surface similar to that
of task 17000, with high AUC values observed over a relatively broad region at small and moderate parameter values.
Overall, the response surfaces show that the influence of $\rho$ and $e$ varies across the Corel5k classification tasks. Several tasks maintain high AUC values over a relatively broad range of parameter settings, whereas others exhibit larger variations across the parameter space. These results indicate that selecting $\rho$ and $e$ through task-specific hyperparameter tuning is appropriate for the proposed XGRVFL-MV model.

\section{Conclusion}
\label{five}

This paper presented XGRVFL-MV, a graph-embedded multi-view Random Vector Functional Link framework for multi-view classification. The proposed model incorporates view-specific RVFL representations, LFDA-based graph embedding, the bounded and asymmetric FleXi Guardian (XG) loss, and a residual-coupling mechanism within a unified optimization framework. The graph regularization preserves the local geometric structure of each view, while the residual-coupling term encourages consistency between view-specific prediction residuals. The resulting optimization problem is solved using an inversion-free first-order optimization procedure based on Nesterov accelerated gradient descent. The proposed model is evaluated on UCI, KEEL, AwA, and Corel5k benchmark datasets and compared with several multi-view RVFL and SVM-based methods. Across the benchmark datasets, XGRVFL-MV achieved the highest average AUC and the lowest average rank among the compared methods. The Friedman, Wilcoxon signed-rank, and pairwise win-tie-loss analyses provided additional evidence for comparing the performance of the proposed model with the baseline methods. The hyperparameter sensitivity analysis further showed that the influence of the regularization parameters varied across datasets and tasks, indicating the importance of dataset-specific parameter selection.
% The present study has several limitations. The experimental evaluation is limited to two-view binary classification problems, and the proposed framework has not been investigated for multi-class classification, regression, or scenarios involving more than two views. In addition, the hyperparameters were selected using a predefined search space, and the sensitivity analysis was conducted on representative datasets and tasks rather than the complete benchmark collection.
Future work will focus on investigating alternative graph construction strategies, evaluating additional loss functions such as the pinball quantile loss, and extending the proposed framework to multi-class classification tasks.
\bibliographystyle{elsarticle-num}
\bibliography{bibfile}

@article{xu2021understanding,
  title={Understanding graph embedding methods and their applications},
  author={Xu, Mengjia},
  journal={SIAM Review},
  volume={63},
  number={4},
  pages={825--853},
  year={2021},
  publisher={SIAM}
}

@article{sugiyama2007dimensionality,
  title={Dimensionality reduction of multimodal labeled data by local fisher discriminant analysis.},
  author={Sugiyama, Masashi},
  journal={Journal of Machine Learning Research},
  volume={8},
  number={5},
  year={2007}
}

@article{yan2006graph,
  title={Graph embedding and extensions: A general framework for dimensionality reduction},
  author={Yan, Shuicheng and Xu, Dong and Zhang, Benyu and Zhang, Hong-Jiang and Yang, Qiang and Lin, Stephen},
  journal={IEEE transactions on pattern analysis and machine intelligence},
  volume={29},
  number={1},
  pages={40--51},
  year={2006},
  publisher={IEEE}
}

@inproceedings{sajid2024wave,
  title={{Wave-RVFL}: A randomized neural network based on wave loss function},
  author={Sajid, M and Quadir, Abdul and Tanveer, Muhammad},
  booktitle={International Conference on Neural Information Processing},
  pages={242--257},
  year={2024},
  organization={Springer}
}

@article{quadir2025enhancing,
  title={Enhancing multiview synergy: Robust learning by exploiting the wave loss function with consensus and complementarity principles},
  author={Quadir, Abdul and Akhtar, Mushir and Tanveer, Muhammad},
  journal={Neural Networks},
  volume={188},
  pages={107433},
  year={2025},
  publisher={Elsevier}
}

@article{arora2026robust,
  title={A robust multi-view support vector machine with the RoBoSS loss function},
  author={Arora, Yash and Gupta, SK and Tanveer, M},
  journal={Neural Networks},
  pages={108937},
  year={2026},
  publisher={Elsevier}
}

@inproceedings{zhou2021asymmetric,
  title={Asymmetric loss functions for learning with noisy labels},
  author={Zhou, Xiong and Liu, Xianming and Jiang, Junjun and Gao, Xin and Ji, Xiangyang},
  booktitle={International Conference on Machine Learning},
  pages={12846--12856},
  year={2021},
  organization={PMLR}
}

@inproceedings{akhtar2024advancing,
  title={Advancing RVFL networks: Robust classification with the HawkEye loss function},
  author={Akhtar, Mushir and Mishra, Ritik and Tanveer, Muhammad and Arshad, Mohd},
  booktitle={International Conference on Neural Information Processing},
  pages={226--240},
  year={2024},
  organization={Springer}
}

@article{zhang2024bounded,
  title={Bounded quantile loss for robust support vector machines-based classification and regression},
  author={Zhang, Jiaqi and Yang, Hu},
  journal={Expert Systems with Applications},
  volume={242},
  pages={122759},
  year={2024},
  publisher={Elsevier}
}

@article{fu2023robust,
  title={Robust regression under the general framework of bounded loss functions},
  author={Fu, Saiji and Tian, Yingjie and Tang, Long},
  journal={European Journal of Operational Research},
  volume={310},
  number={3},
  pages={1325--1339},
  year={2023},
  publisher={Elsevier}
}

@article{quadir2024multiview,
  title={Multiview random vector functional link network for predicting DNA-binding proteins},
  author={Quadir, Abdul and Sajid, M and Tanveer, Muhammad},
  journal={arXiv preprint arXiv:2409.02588},
  year={2024}
}

@article{akhtar2025towards,
  title={Towards robust and inversion-free randomized neural networks: The XG-RVFL framework},
  author={Akhtar, Mushir and Kumari, A and Sajid, M and Quadir, A and Arshad, Mohd and Suganthan, PN and Tanveer, M},
  journal={Pattern Recognition},
  pages={112711},
  year={2025},
  publisher={Elsevier}
}

@article{malik2023random,
  title={Random vector functional link network: Recent developments, applications, and future directions},
  author={Malik, Ashwani Kumar and Gao, Ruobin and Ganaie, MA and Tanveer, Muhammad and Suganthan, Ponnuthurai Nagaratnam},
  journal={Applied Soft Computing},
  volume={143},
  pages={110377},
  year={2023},
  publisher={Elsevier}
}

@article{belkin2006manifold,
  title={Manifold regularization: A geometric framework for learning from labeled and unlabeled examples.},
  author={Belkin, Mikhail and Niyogi, Partha and Sindhwani, Vikas},
  journal={Journal of machine learning research},
  volume={7},
  number={11},
  year={2006}
}

@article{tanveer2025grvfl,
  title={{GRVFL-MV}: Graph random vector functional link based on multi-view learning},
  author={Tanveer, Muhammad and Sharma, RK and Sajid, M and Quadir, Abdul},
  journal={Information Sciences},
  volume={704},
  pages={121947},
  year={2025},
  publisher={Elsevier}
}

@incollection{yang2009artificial,
  title={Artificial neural networks},
  author={Yang, Xiaojun},
  booktitle={Handbook of research on geoinformatics},
  pages={122--128},
  year={2009},
  publisher={IGI Global Scientific Publishing}
}

@article{quadir2025randomized,
  title={Randomized based restricted kernel machine for hyperspectral image classification},
  author={Quadir, Abdul and Tanveer, M},
  journal={arXiv preprint arXiv:2503.05837},
  year={2025}
}

@article{quadir2025trkm,
  title={TRKM: Twin restricted kernel machines for classification and regression},
  author={Quadir, Abdul and Tanveer, Muhammad},
  journal={Neural Networks},
  pages={108449},
  year={2025},
  publisher={Elsevier}
}

@article{sajid2024neuro,
  title={Neuro-fuzzy random vector functional link neural network for classification and regression problems},
  author={Sajid, M and Malik, Ashwani Kumar and Tanveer, Muhammad and Suganthan, Ponnuthurai N},
  journal={IEEE Transactions on Fuzzy Systems},
  volume={32},
  number={5},
  pages={2738--2749},
  year={2024},
  publisher={IEEE}
}

@article{malik2022extended,
  title={Extended features based random vector functional link network for classification problem},
  author={Malik, Ashwani Kumar and Ganaie, MA and Tanveer, M and Suganthan, Ponnuthurai N},
  journal={IEEE Transactions on Computational Social Systems},
  volume={11},
  number={4},
  pages={4744--4753},
  year={2022},
  publisher={IEEE}
}

@article{qin2025survey,
  title={A survey on representation learning for multi-view data},
  author={Qin, Yalan and Zhang, Xinpeng and Yu, Shui and Feng, Guorui},
  journal={Neural Networks},
  volume={181},
  pages={106842},
  year={2025},
  publisher={Elsevier}
}

@article{liu2025multi,
  title={Multi-view structural twin support vector machine with the consensus and complementarity principles and its safe screening rules},
  author={Liu, Qianfei and Chen, Chen and Huang, Ting and Meng, Yan and Wang, Huiru},
  journal={Expert Systems with Applications},
  volume={265},
  pages={125814},
  year={2025},
  publisher={Elsevier}
}

@article{yan2025multi,
  title={Multi-view learning with enhanced multi-weight vector projection support vector machine},
  author={Yan, Xin and Wang, Shuaixing and Chen, Huina and Zhu, Hongmiao},
  journal={Neural Networks},
  volume={185},
  pages={107180},
  year={2025},
  publisher={Elsevier}
}

@article{hu2024multiview,
  title={Multiview large margin distribution machine},
  author={Hu, Kun and Xiao, Yingyuan and Zheng, Wenguang and Zhu, Wenxin and Hsu, Ching-Hsien},
  journal={IEEE Transactions on Neural Networks and Learning Systems},
  volume={36},
  number={2},
  pages={2395--2409},
  year={2024},
  publisher={IEEE}
}

@article{chen2025multi,
  title={Multi-view support vector machine classifier via {L0/1} soft-margin loss with structural information},
  author={Chen, Chen and Liu, Qianfei and Xu, Renpeng and Zhang, Ying and Wang, Huiru and Yu, Qingmin},
  journal={Information Fusion},
  volume={115},
  pages={102733},
  year={2025},
  publisher={Elsevier}
}

@article{he2025new,
  title={A new multi-view support vector machine with v-property},
  author={He, Xueyang and Pang, Xinying and Ji, Zhijian},
  journal={Expert Systems with Applications},
  pages={130097},
  year={2025},
  publisher={Elsevier}
}

@misc{asuncion2007uci,
  title={{UCI} machine learning repository},
  author={Asuncion, Arthur and Newman, David and others},
  year={2007},
  publisher={Irvine, CA, USA}
}

@article{lampert2013attribute,
  title={Attribute-based classification for zero-shot visual object categorization},
  author={Lampert, Christoph H and Nickisch, Hannes and Harmeling, Stefan},
  journal={IEEE Transactions on Pattern Analysis and Machine Intelligence},
  volume={36},
  number={3},
  pages={453--465},
  year={2013},
  publisher={IEEE}
}

@article{derrac2015keel,
  title={Keel data-mining software tool: Data set repository, integration of algorithms and experimental analysis framework},
  author={Derrac, J and Garcia, S and Sanchez, L and Herrera, F},
  journal={J. Mult. Valued Logic Soft Comput},
  volume={17},
  pages={255--287},
  year={2015}
}

@article{friedman1940comparison,
  title={A comparison of alternative tests of significance for the problem of m rankings},
  author={Friedman, Milton},
  journal={The Annals of Mathematical Statistics},
  volume={11},
  number={1},
  pages={86--92},
  year={1940},
  publisher={JSTOR}
}

@article{demvsar2006statistical,
  title={Statistical comparisons of classifiers over multiple data sets},
  author={Dem{\v{s}}ar, Janez},
  journal={The Journal of Machine Learning Research},
  volume={7},
  pages={1--30},
  year={2006},
  publisher={JMLR. org}
}

@article{wang2023safe,
  title={Safe screening rules for multi-view support vector machines},
  author={Wang, Huiru and Zhu, Jiayi and Zhang, Siyuan},
  journal={Neural Networks},
  volume={166},
  pages={326--343},
  year={2023},
  publisher={Elsevier}
}

@article{datta2008image,
  title={Image retrieval: Ideas, influences, and trends of the new age},
  author={Datta, Ritendra and Joshi, Dhiraj and Li, Jia and Wang, James Z},
  journal={ACM Computing Surveys (Csur)},
  volume={40},
  number={2},
  pages={1--60},
  year={2008},
  publisher={ACM New York, NY, USA}
}

@article{eidenberger2004statistical,
  title={Statistical analysis of content-based {MPEG-7} descriptors for image retrieval},
  author={Eidenberger, Horst},
  journal={Multimedia Systems},
  volume={10},
  number={2},
  pages={84--97},
  year={2004},
  publisher={Springer}
}

@article{jia2025deep,
  title={Deep Multi-view Least Squares Support Vector Machine with Consistency and Complementarity Principle based on Cross-Output Knowledge Transfer},
  author={Jia, Shuangrui and Liang, Sijie and Mo, Ziyi and Liu, Chunxiao and Wang, Huiru and Chen, Chen},
  journal={Expert Systems with Applications},
  pages={129406},
  year={2025},
  publisher={Elsevier}
}

@article{xie2023deep,
  title={Deep multi-view multiclass twin support vector machines},
  author={Xie, Xijiong and Li, Yanfeng and Sun, Shiliang},
  journal={Information Fusion},
  volume={91},
  pages={80--92},
  year={2023},
  publisher={Elsevier}
}

@article{lou2024multi,
  title={Multi-view universum support vector machines with insensitive pinball loss},
  author={Lou, Chunling and Xie, Xijiong},
  journal={Expert Systems with Applications},
  volume={248},
  pages={123480},
  year={2024},
  publisher={Elsevier}
}

@article{xie2015multi,
  title={Multi-view twin support vector machines},
  author={Xie, Xijiong and Sun, Shiliang},
  journal={Intelligent Data Analysis},
  volume={19},
  number={4},
  pages={701--712},
  year={2015},
  publisher={SAGE Publications Sage UK: London, England}
}

@article{berahmand2025comprehensive,
  title={A comprehensive survey on multi-view classification: Methods, applications, and challenges},
  author={Berahmand, Kamal and Daneshfar, Fatemeh and Rahmaninia, Maryam and Haghighat, Maryam and Jalili, Mahdi},
  journal={ACM Transactions on Intelligent Systems and Technology},
  volume={16},
  number={6},
  pages={1--34},
  year={2025},
  publisher={ACM New York, NY}
}

@article{xu2013survey,
  title={A survey on multi-view learning},
  author={Xu, Chang and Tao, Dacheng and Xu, Chao},
  journal={arXiv preprint arXiv:1304.5634},
  year={2013}
}

@article{jacobs1988increased,
  title={Increased rates of convergence through learning rate adaptation},
  author={Jacobs, Robert A},
  journal={Neural Networks},
  volume={1},
  number={4},
  pages={295--307},
  year={1988},
  publisher={Elsevier}
}

@article{gori1992problem,
  title={On the problem of local minima in backpropagation},
  author={Gori, Marco and Tesi, Alberto},
  journal={{IEEE Transactions on Pattern Analysis and Machine Intelligence}},
  volume={14},
  number={1},
  pages={76--86},
  year={1992}
}

@article{pao1994learning,
  title={Learning and generalization characteristics of the random vector functional-link net},
  author={Pao, Yoh-Han and Park, Gwang-Hoon and Sobajic, Dejan J},
  journal={Neurocomputing},
  volume={6},
  number={2},
  pages={163--180},
  year={1994},
  publisher={Elsevier}
}

@article{suganthan2021origins,
  title={On the origins of randomization-based feedforward neural networks},
  author={Suganthan, Ponnuthurai N and Katuwal, Rakesh},
  journal={Applied Soft Computing},
  volume={105},
  pages={107239},
  year={2021},
  publisher={Elsevier}
}

@article{zhang2016survey,
  title={A survey of randomized algorithms for training neural networks},
  author={Zhang, Le and Suganthan, Ponnuthurai N},
  journal={Information Sciences},
  volume={364},
  pages={146--155},
  year={2016},
  publisher={Elsevier}
}

@article{huang2006extreme,
  title={Extreme learning machine: theory and applications},
  author={Huang, Guang-Bin and Zhu, Qin-Yu and Siew, Chee-Kheong},
  journal={Neurocomputing},
  volume={70},
  number={1-3},
  pages={489--501},
  year={2006},
  publisher={Elsevier}
}

\end{document}